\documentclass{article}
\usepackage[a4paper,left=1in, right=1in, top=1in, bottom=1in]{geometry}

\usepackage{natbib}% Citation support using natbib.sty
\bibpunct[, ]{(}{)}{;}{a}{}{,}% Citation support using natbib.sty
\renewcommand\bibfont{\fontsize{10}{12}\selectfont}% Bibliography support using natbib.sty

\usepackage{amsmath}                       %for equations
\usepackage{amsthm}                        %theorem environments
\usepackage{multirow}                      %for tables
\usepackage{booktabs}                      %for tables
\usepackage{color,soul}                    %for highlighting text
\usepackage{caption}                       %for table captions
\usepackage[hyphens]{url}                  %to split long u\gls{RL} also at -
\usepackage{subfiles}                      %to add the rebuttal at the end of the paper
\usepackage{setspace}                      %to obtain a good table size
\usepackage{arydshln}	                   %defines dashed rules in tables
\usepackage{amsfonts}  
\usepackage{tikz,pgfplots}
\pgfplotsset{compat=1.18}
\usepackage{caption}
\usepackage{subcaption}
\usetikzlibrary{patterns}
\usepackage{booktabs}  
\usepackage{multirow}
\usepackage{rotating}

\usepackage{amsthm}
 \newtheorem{observation}{Observation}
 \usepackage{graphicx}

\newcommand{\NN}{N}

\newcommand{\K}{\mathcal{K}}
\newcommand{\s}{\mathbf{s}}

\usepackage{accents}

\usepackage{siunitx}
 \usepgfplotslibrary{statistics}
 \usepackage{adjustbox}
 \usepgfplotslibrary{fillbetween}
 \usepackage[acronym,nomain,nonumberlist, section=subsection]{glossaries}
 \usetikzlibrary{arrows.meta,
                chains,
                positioning,
                shapes.geometric,
                fit
                }

\usepackage[linesnumbered,ruled, vlined]{algorithm2e}    %for algorithm
\usepgfplotslibrary{fillbetween}

\usepackage{csquotes}

\begin{document}
\newacronym{drl}{DRL}{deep reinforcement learning}
\newacronym{PPO}{PPO}{Proximal Policy Optimization}
\newacronym{RL}{RL}{reinforcement learning}
\newacronym{MDPs}{MDPs}{Markov decision processes}
\newacronym{MDP}{MDP}{Markov decision process}
\newacronym{AM}{AM}{additive manufacturing}
\newacronym{CM}{CM}{conventional manufacturing}
\newacronym{AVI}{AVI}{Approximate value iteration}
\newacronym{DCL}{DCL}{Deep controlled learning}
\newacronym{EPL}{EPL}{Endogenously parameterised learning}
\newacronym{VFA}{VFA}{Value function approximation}
\newacronym{PFA}{PFA}{Policy function approximation}
\newacronym{IWA}{IWA}{Iterative weight adjustment}
\newacronym{BSP}{BSP}{Base stock policy}
\newacronym{EOQ}{EOQ}{Economic order quantity}

\title{Solving Dual Sourcing Problems with Supply Mode Dependent Failure Rates}

\author{Fabian Akkerman - University of Twente \cr Nils Knofius - Fieldmade AS \cr Matthieu van der Heijden - University of Twente \cr Martijn Mes - University of Twente}
\date{}

\maketitle

\begin{abstract}
This paper investigates dual sourcing problems with supply mode dependent failure rates, particularly relevant in managing spare parts for downtime-critical assets. To enhance resilience, businesses increasingly adopt dual sourcing strategies using both conventional and additive manufacturing techniques. This paper explores how these strategies can optimise sourcing by addressing variations in part properties and failure rates. A significant challenge is the distinct failure characteristics of parts produced by these methods, which influence future demand. To tackle this, we propose a new iterative heuristic and several reinforcement learning techniques combined with an endogenous parameterised learning (EPL) approach. This EPL approach -- compatible with any learning method -- allows a single policy to handle various input parameters for multiple items. In a stylised setting, our best policy achieves an average optimality gap of 0.4\%. In a case study within the energy sector, our policies outperform the baseline in 91.1\% of instances, yielding average cost savings up to 22.6\%.
\end{abstract}

% \begin{keywords}
% Inventory management; Dual sourcing; Reinforcement learning; Spare parts; Additive manufacturing
% \end{keywords}

Unplanned asset downtime is a costly issue in many industries, as over the past five years, the average cost of downtime across the consumer goods, automotive, oil and gas, and heavy industry sectors has doubled, rising from $\$400{\small,}000$ to $\$800{\small,}000$ per hour \citep{siemens2024}. For example, offshore oil and gas companies face annual costs of $\$49$ million due to unplanned downtime \citep{ge2016}, while the aerospace industry loses $\$6$ billion annually because of grounded aircraft \citep{airotime2023}. Globally, the 500 largest companies are estimated to lose $\$1.4$ trillion per year due to unplanned downtime, equivalent to $11\%$ of their revenues. Downtime results in lost sales, production delays, safety hazards, and significant expenses for (temporary) fixes. A major cause of unplanned downtime is the failure of critical parts (such as those found in manufacturing equipment), which accounts for $45\%$ of reported downtime instances \citep{vanson2023}. {The financial impact of unplanned downtime highlights the need for more efficient management strategies. This paper proposes a novel approach building on optimisation and machine learning techniques to address these challenges.}

Spare parts' repair and replenishment lead times are lengthy across various industries. For instance, in the aerospace sector, the average replenishment lead time of spare parts is approximately 100 days \citep{deloitte2024}. For some parts used on offshore oil and gas platforms, lead times can extend up to 16 weeks \citep{Everett2022}, while in the automotive industry, it can take up to 9 months \citep{antich2023}. Traditionally, downtime of critical assets is mitigated by stocking spare parts close to the installed base. However, since these bases are often geographically dispersed, a substantial portion of the inventory must be stocked at multiple global locations. Additionally, complex assets contain thousands of expensive components, making the stocking of critical spare parts a costly endeavor.

To improve the balance between asset downtime and spare parts inventory investment, reducing lead times is crucial. One approach to achieve this is to (partially) replace the sourcing of spare parts from \gls{CM} sources by \gls{AM}, commonly referred to as 3D printing \citep{mecheter2022}. \gls{AM} is a rapidly growing printing technology. The global AM market was valued at $\$22$ billion in 2023 and is expected to grow by $23\%$ by 2032 \citep{fortune2023}. In theory, \gls{AM} can enable print-on-demand, reducing the need for inventories \citep{Muir2018AdditiveChain}. However, {in reality,} the total time from identifying the need for a spare part to its availability can still take weeks due to post-processing, testing, qualifying, and approving parts {\citep{Knofius2020ImprovingOption, PERON2022}}, which is unacceptable for downtime-prone assets. \gls{AM} and \gls{CM} parts differ in lead times, costs, and failure behavior. \gls{AM} parts may fail earlier due to production characteristics like porosity, but they can also have superior designs, such as single-piece prints that eliminate weaknesses from assembling multiple subcomponents \citep{Lindemann2015TowardsManufacturing,Wits2016HowStrategies}.

The choice between either \gls{AM} or \gls{CM} sourcing is straightforward when one manufacturing technique dominates, offering shorter lead times, lower manufacturing costs, and a lower failure rate. However, this dominance is not always evident, necessitating a trade-off. Methods supporting the trade-off between AM and CM are available in the literature, see, e.g., \citet{WESTERWEEL2018} and \citet{Song2016StockLogistics}. Nonetheless, relying on a single sourcing option may not always be optimal. A dual sourcing approach can leverage the strengths of both methods, e.g., using CM for the bulk of spare parts demand and AM to cover uncertain peaks in demand. At first glance, this resembles a standard dual sourcing model \citep{SVOBODA20211, xinvanmiegem2023}. However, a crucial problem aspect is missing in these standard dual sourcing models: the significant complication arising from the difference in failure rates between \gls{CM} and \gls{AM}, which affects future spare parts demand. Ordering and installing an AM item with a higher failure rate increases expected future demand, potentially offsetting the initial benefits of shorter lead times.

We consider the challenge of managing the spare parts inventory for critical components installed in an asset prone to downtime. When a part fails, it is replaced by an on-hand spare part -- a stock keeping unit (SKU) -- or it is backordered if no spare part is available. Our task is to decide on the dual sourcing of spare part SKUs from an AM or a CM source. Once a part is installed, it begins to wear and can fail with a certain predefined probability. Due to production differences, AM and CM parts have different failure probabilities. Thus, the expected number of failures depends on the proportion of AM and CM parts installed in the asset. This complex problem, where the demand is endogenously dependent on the proportion of operating AM and CM items, has been addressed in the literature only by \citet{Knofius2020ImprovingOption}, who propose an exact dual sourcing approach applicable only to small-scale problem instances.

In this paper, we provide multiple novel approaches to solve this dual sourcing problem for realistically sized problem instances {of up to $150$ parts in the installed base}. First, we propose an iterative heuristic method that can be combined with most dual sourcing solution procedures, finding a policy in seconds. Second, we develop several \gls{RL} methods. We consider RL policies from the three main classes: (i) value-based, (ii) policy-based, and (iii) actor-critic, where the latter is a combination of the former two. These RL approaches' flexibility enables us to relax common assumptions in traditional operational research, such as specific demand distributions (e.g., Poisson, compound Poisson) and deterministic lead times. However, the flexibility of RL comes at the cost of significant training time for each spare part type. Given that companies may manage thousands of critical spare parts, training separate policies for each is impractical, especially since policies need to be retrained when product characteristics change. To overcome this challenge, we propose a novel approach, called Endogenously Parameterised Learning (EPL), for which we train a single policy on a grid of critical parameter settings, covering a range of SKUs. This approach allows one trained policy to be applied to a group of SKUs, eliminating the need for thousands of separate policies that need frequent retraining. We demonstrate the use of EPL by embedding the three RL policy classes in it. EPL can serve as a foundation for future RL models in inventory management.

{Our model and solution approach offer the following key advantages: (i) we account for differences in failure behavior between AM and CM parts, enabling a more informed trade-off in deciding when to use AM versus CM in a dual sourcing setting, (ii) our reinforcement learning approach effectively handles large state spaces -- which grows polynomially with the installed base for our problem, see \citet{Knofius2020ImprovingOption} -- making it suitable for a wide range of applications, including scenarios with an installed base of up to 150 parts, and (iii) by training our model on a grid of critical parameter settings, we can quickly identify near-optimal dual sourcing policies and their impacts across a wide range of problem instances within the grid.}

The remainder of this paper is structured as follows. Section~\ref{sec:litreview} positions our work within the literature, and clarifies our contribution. Section~\ref{sec:prob} defines our dual sourcing problem in detail and models it as an \gls{MDP}. Section~\ref{sec:Approach} details our heuristic method, our RL approach, and a novel approach for learning a single policy for multiple SKUs. In Section~\ref{sec:Experiments}, we present numerical experiments. {We first study a small synthetic case of $10$ instances with the purpose to benchmark our proposed policies against an exact policy, after which we move to a real-world case in the energy sector -- which includes $1215$ large instances -- with the purpose to distill managerial insights.} Finally, Section~\ref{sec:Conclusion} summarises our findings and suggests directions for future research.

\section{Literature Review}
\label{sec:litreview}
Our research is related to three areas in the literature that we will discuss, namely \gls{AM} for spare parts (Section~\ref{subsec:litreviewAM}), dual sourcing (Section~\ref{subsec:litreviewDual}), and \gls{RL} for inventory problems (Section~\ref{litrev:RL}). In Section~\ref{subsec:Contribution}, we state our contribution to the existing body of literature. 

\subsection{\gls{AM} for Spare Parts}
\label{subsec:litreviewAM}

The possible application of \gls{AM} for spare parts supply chains have attracted considerable attention in the literature in the last years, as \gls{AM} is in particular useful for one piece manufacturing of customised products. Most spare parts assortments for high-tech assets contain many slow movers with conventionally long lead times, for which shorter lead times are highly desirable. Early papers on the application of \gls{AM} to spare parts focused on the selection of parts that are suitable for \gls{AM} \citep{Lindemann2015TowardsManufacturing, Knofius2016}. These papers proceed from the characteristics of parts that are currently sourced through \gls{CM}. An aspect that is typically challenging to address is the difference in failure behaviour. \citet{Wits2016HowStrategies} explain that part properties will differ significantly between \gls{AM} and \gls{CM} produced items, leading to differences in failure behaviour and thus impacting spare parts demand. 

Numerous model variants have been investigated. \citet{WESTERWEEL2018} study the trade-off between deploying AM or CM for spare parts.  \citet{Sgarbossa2021ConventionalDemand} include multiple \gls{AM} and \gls{CM} techniques to decide between \gls{AM} and \gls{CM}. \citet{Song2016StockLogistics} examine the trade-off between \gls{AM} and \gls{CM} across multiple SKUs, considering the finite capacity of 3D printers.  

\citet{KNOFIUS2019} analyse the impact of using \gls{AM} print parts in one piece (consolidation) versus assembling them from multiple subcomponents on life cycle costs. They find that initial cost savings from \gls{AM} may be offset by higher maintenance costs if expensive consolidated \gls{AM} parts are needed for repairs, instead of the cheaper subcomponents in \gls{CM}. Additionally, \citet{Cantani2024} demonstrate that AM, by enabling decentralised manufacturing, influences the spare parts supply chain design. 

\citet{Westerweel2020PrintingManufacturing} show the value of \gls{AM} for decentralised spare parts production, especially for remote assets like those in defence missions. AM remains viable even if the part quality is lower than centrally produced spare parts. Further research has revealed that \gls{AM} of spare parts is useful in cases of supply disruptions, such as natural disasters, unexpected demand surges, or when a part supplier discontinues production during the asset's life cycle (e.g., \citet{Ivanov2019TheAnalytics}). For a more comprehensive discussion, we refer to the literature reviews by \citet{Verboeket2019TheAgenda}, and \citet{Kunovjanek2020AdditiveReview}.

\subsection{Dual Sourcing}
\label{subsec:litreviewDual}
Dual sourcing models typically differentiate between two supply options: one that is inexpensive but has a long resupply lead time (regular supply), and another that is expensive but offers a shorter resupply lead time (expedited order). The first contribution to the dual sourcing literature is by \citet{Barankin1961ACase}, who discuss a single-period model with emergency shipments. Since then, numerous papers have emerged on dual sourcing. 

\citet{Scheller-Wolf2006InventoryPolicies} propose a heuristic single index policy that monitors the inventory position over the long (regular) lead time. They use two order-up-to levels: one for expedited orders and one for regular orders. First, an expedited order is placed to raise the inventory to the expedited base stock level if needed. Then, a regular order is placed to reach the higher regular base stock level. \citet{Veeraraghavan2008NowSystems} refine this into a dual index policy, monitoring two inventory positions: one covering the replenishment orders scheduled to arrive within the expedited lead time only, and one covering all outstanding replenishment orders. Expedited orders are based on the first inventory position, followed by regular orders based on the second. The dual index policy appears to be close to the optimal solution found via dynamic programming. As dynamic programming suffers from the curse of dimensionality, we will deploy the dual index policy as a building block for our heuristic approach in Section~\ref{sec:Approach}. {\citet{sun2019} propose a robust dual sourcing inventory model that minimises total costs under uncertainty, offering a capped dual index policy to mitigate order variability. Their results demonstrate the policy’s optimality in nonstationary demand scenarios, providing both cost-effectiveness and practical applicability through closed-form solutions. This capped policy differs from prior dual index policies by incorporating constraints on slow order quantities and smoothing orders. In \citet{XIONG2022}, robust optimisation techniques are integrated with a data-driven approach to handle purchase price and demand uncertainties for dual sourcing. Using historical data, they construct uncertainty sets to optimise inventory policies dynamically, which outperforms existing dual index benchmarks.}

Regarding dual sourcing with \gls{AM} and CM, \citet{Song2016StockLogistics} explore the utilisation of \gls{AM} as an emergency channel capable of producing multiple spare parts on-demand. In their model, \gls{AM} equipment is capacitated (modelled as an M/D/1 queue), offering faster but more expensive resupply compared to \gls{CM}. However, they note that the utilisation of \gls{AM} equipment tends to remain low. They assume that parts manufactured through \gls{AM} exhibit the same failure behaviour as those produced conventionally. {In \citet{VANOERS2024} however, the authors demonstrate that the deployment of high-speed metal AM in downstream echelons can significantly reduce costs, inventory levels, and dependency on upstream stockouts, while also improving supply chain resilience. Their analysis highlights that AM’s integration into dual sourcing strategies reduces overall system sensitivity to supply uncertainty, making it a robust option for mission-critical applications in military and humanitarian operations.}

To date, the only papers we are aware of that address the relation between sourcing decisions and demand are \citet{Knofius2020ImprovingOption} and \citet{PERON2022}. The former approach, based on \gls{MDP}s, is limited to relatively small problem instances due to the curse of dimensionality, {i.e., the state space grows polynomially with the installed base size and the maximum number of spare parts circulating in the system, and can therefore not scale to instance sizes with $\gg 50$ items in the installed base.} Our aim in this paper is to develop new methods applicable to {such} problem instances. \citet{PERON2022} also consider the effect of sourcing decisions on demand, but do not consider a dual sourcing policy. {The main insights from both works are: (i) dual sourcing can significantly reduce costs, even if the AM part is significantly more expensive and has higher failure rate compared to the CM part, (ii) printing spare parts on demand is not optimal for downtime critical spare parts and stock remains necessary, (iii) in most cases, AM serves as an emergency source when a large portion of the CM base stock is depleted, (iv) dual sourcing is most promising if single sourcing of CM parts leads to high backorder and holding costs, and (v) the failure rate of AM parts -- relative to the CM part failure rate -- is one of the main determinants of the adoption of AM \citep{Knofius2020ImprovingOption, PERON2022}.}

Given the advancements in computational power and algorithms, a promising {and scalable} alternative {to exact approaches} is a general solution framework such as \gls{RL}, as suggested by \citet{SVOBODA20211} in their comprehensive review of dual sourcing literature, and by \citet{xinvanmiegem2023}.

\subsection{\gls{RL} for Inventory Problems}\label{litrev:RL}
An early paper on \gls{RL} for inventory management is \citet{Gijsbrechts2018CanProblems}. They use a deep \gls{RL} approach (employing neural networks) and find that their general solution framework is matching or even outperforming commonly used heuristics. However, they also acknowledge that these results were achieved with substantial tuning effort. For the dual sourcing problem they consider, it is not yet clear whether an RL framework or a dedicated heuristic is the better approach. 

\citet{temizoz2023deep} introduce a novel deep \gls{RL} framework based on approximate policy iteration that is specifically designed to solve inventory problems. The authors emphasise a significant advantage: a single set of hyperparameters proves effective across all their experiments. This implies high robustness and reduces the necessity for fine-tuning hyperparameters to individual problem instances. Therefore, this technique emerges as a promising candidate for application to our specific dual sourcing problem.

\citet{tang2023} apply online learning to dual sourcing problems with periodic review to learn unknown demand distributions from demand realisations. They use a dual index policy as a heuristic for sourcing decisions, following \citet{Veeraraghavan2008NowSystems}. Their focus on demand learning is distinctly different from our approach, as we explore broader RL applications and novel heuristic methods to enhance dual sourcing strategies.

\citet{botcherRNN2023} explore dual sourcing inventory systems using recurrent neural networks (RNNs) to learn near-optimal inventory policies. While their approach offers valuable insights, it may not be ideal for our context. Our low demand rate and long planning horizons could lead to inefficiencies, given that their method's computational requirements scale linearly with the length of the time horizon. Additionally, RNNs potentially face challenges with long sequences and sparse meaningful events, which is a concern in our low-demand scenario. 
 
In their review, \citet{BOUTE2022401} highlight several challenges in applying RL to inventory control, with a key issue being the development of robust RL policies. RL is often unstable and non-robust, as also noted by \citet{lutjens2020}, meaning that trained policies fail to generalise when problem parameters change. This necessitates frequent retraining for varying parameters and individual SKUs, which is impractical for inventory management involving thousands of spare parts. Although offline training is computationally intensive, it is often overlooked because the application of policies in real-time (online) is very fast. Some recent works address this challenge using supervised machine learning methods, building on weighted sample average approximation (wSAA) to find a \enquote*{global learning} model that exploits cross-learning effects. This approach enables the use of data from all products simultaneously for finding better forecasts, supporting affine or exact (MIP-based) reordering policies, see \citet{lin2022} and \citet{schmidt2023}. {Similarly, \citet{temizoz2024} introduce a Super-Markov Decision Process (Super-MDP) within their Train, then Estimate and Decide (TED) framework, which enables robust deployment to unseen instances without retraining. Their RL agent achieves strong generalisation performance in dynamic inventory environments by dynamically adapting policies through real-time parameter estimation.}

To address robustness, approaches outside inventory control, such as incorporating state and action noise with adversarial policies \citep{pinto2017, Gleave2020Adversarial}, have been proposed. However, integrating such techniques into inventory control is not straightforward. To the best of our knowledge, there is no existing method for training robust RL policies specifically for inventory control. Our work aims to fill this gap by developing a robust RL approach that can handle parameter variations and reduce the need for frequent retraining, making it viable for large-scale inventory management. Compared to \citet{lin2022} and \citet{schmidt2023}, our focus is on leveraging deep neural networks trained using RL techniques to obtain a reordering policy directly, which allows for better scalability across different SKUs, i.e., scaling from slow to fast movers.

\subsection{Contribution}
\label{subsec:Contribution}

The literature shows a gap in addressing dual sourcing problems with sourcing decisions influencing future demand, lacking scalable policies for realistically sized problems. Despite growing interest in RL, it proves non-robust for spare parts management with changing parameters, leading to ineffective policies for different items. This reveals a need for scalable and generalizable solutions. Our contribution to the scientific body of literature includes:
\begin{enumerate}
    \item We design three distinct classes of \gls{RL} approaches to obtain policies for the same problem, each trained separately on individual SKUs, and evaluate the conditions under which each method is preferred.
    \item We introduce an approach where a single \gls{RL} policy is trained to generalise across multiple SKUs. We evaluate this approach on trained SKUs and demonstrate its effectiveness on SKUs that are not included in the training set.
    \item We develop a novel iterative heuristic for large-scale dual sourcing problems with sourcing-dependent demand, extending beyond slow movers. 
    \item We apply our proposed approaches to a real-world case study in the energy sector, showcasing the practical applicability of our model and methodology.
\end{enumerate}
To facilitate reproducibility, our code is publicly available at: \url{https://github.com/frakkerman/dual_sourcing_am}.

\section{Problem Formulation}
\label{sec:prob}
We formalise the dual sourcing problem as an MDP in this section. Section~\ref{subsec:assumption} provides the problem setting. Section~\ref{subsec:notation} outlines the model notation. Sections~\ref{subsec:State space} and~\ref{subsec:decisio} describe the constraints on the state and decision spaces, respectively. Section~\ref{subsec:transition} covers transitions and transition probabilities. Finally, Section~\ref{subsec:CostFun} explains the evaluation of expected costs for given states and decisions.

\subsection{Problem Setting and Assumptions}
\label{subsec:assumption}

We consider a service provider that orders spare parts for a specific critical component using either \gls{CM} or \gls{AM}. Both supply modes are characterised by different failure rates, replenishment lead time, and piece price. \gls{CM} may require significant setup costs, necessitating batch ordering, in contrast to \gls{AM} where setup costs are typically low once the digital asset for the part is created. Parts are non-repairable and always need to be replaced whenever they fail. The service provider has to decide how many batches to order from the \gls{CM} suppliers, and how may items from the \gls{AM} supplier. 
\medbreak
We formulate the problem as a discrete-time, periodic-review model. The sequence of events at a given period $t$ is as follows:
\begin{enumerate}   
\item At the start of a period, the current state of the system is recorded as input to the reorder policy.
    \item Potential replenishment orders from the CM and/or AM source are placed.
    \item Holding, reordering, maintenance, and backorder costs are incurred.
    \item During a period, parts may fail and are replaced by the stock on-hand, or alternatively backordered.
     \item At the end of a period, possible replenishment arrives, which is put in stock and/or used for backorder clearing without further delay.
\end{enumerate}
During this event sequence, the decision-maker has to decide on the reordering decision at step (2). In addition, we make the following assumptions to obtain a tractable model with an \emph{exact} solution procedure.
\begin{enumerate}
\item \textit{Lead times are deterministic.} 

Resupply lead times are usually part of contractual agreements that guarantee a high delivery reliability within the specified lead time.
\medbreak
\item\textit{Lead times are a multiple of the review period.}

This assumption simplifies the analysis because orders can only arrive at the end of a period. This is an assumption with marginal effect, as in practice we can always select a review period that ensures this.
\medbreak
\item\textit{Demand per item is i.i.d. between subsequent periods, conditional on the number of operating \gls{CM} and \gls{AM} parts at the beginning of the period, and follows a discrete distribution for the individual demand.}

The demand is endogenous, and failures and replacements may change the proportion of operating \gls{CM} and \gls{AM} items during a period. However, demand is independent from the demand in the previous period.
\medbreak
\item \textit{{CM parts are ordered in fixed batch sizes.}}

{This assumption ensures tractability for the exact policy, other policies do not require it. Our formulation permits a batch size of one with multiple batches per time step, effectively allowing variable batch sizes.}
\medbreak
\item \textit{Items with lower failure rate are installed first, if any.}

\citet{Knofius2020ImprovingOption} observe that usually items with lower failure rate are installed first. Without loss of generality, we follow this result and assume that throughout this paper.
\medbreak
\item \textit{The total number of items circulating in the inventory system is bounded.}

Although different failure rates for \gls{AM} and \gls{CM} parts could affect the optimal inventory level, we assume it remains relatively stable regardless of the part composition. Therefore, we use the average best-performing number of circulating items, determined through a greedy approach.
\end{enumerate}

Together, these assumptions allow us to develop a tractable model suitable for an exact solution. While our proposed \gls{RL} approach only requires a bounded decision space (Assumption~6), we use all the assumptions when applying RL. This ensures we can validate and fairly compare the RL policies against the exact benchmark.

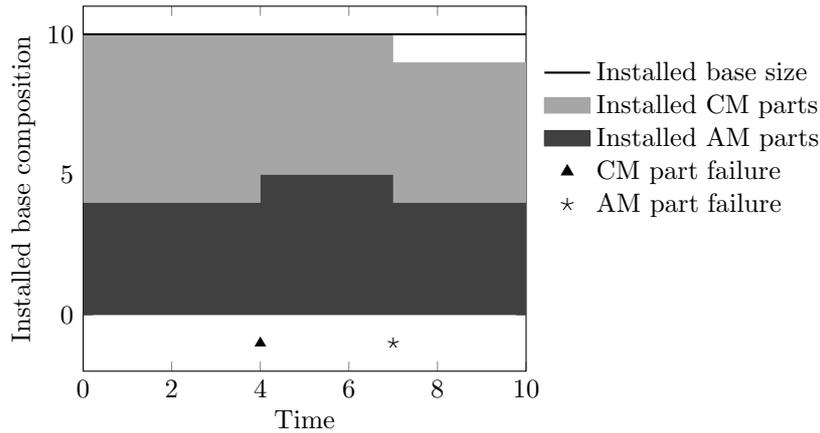
\begin{figure}[hbtp]
    \centering
  \begin{tikzpicture}[scale = 0.85]
            \begin{axis}[
                legend style={at={(1.2,0.75)},anchor=west, draw=none},
                legend columns=1,
                legend cell align={left},
               ytick distance = 5,
                ymin = -2 ,ymax=11,
                xmin=0, xmax=10,
                 xlabel = Time,
                xlabel near ticks,
                 ylabel near ticks,
                ylabel= Installed base composition
                        ]

                 \addplot[thick] %
                coordinates {
                   (0,10)
                         (14,10)

                };

            \addplot[draw=none,name path = B, forget plot] %
                coordinates {
                   (0,4)
                   (4,4)
                   (4,5)
                   (7,5)
                   (7,4)
                   (14,4)

                };

            \addplot[draw=none,name path = A, forget plot] %
                coordinates {
                   (0,10)
                   (7,10)
                   (7,9)
                   (14,9)
                };

\addplot [black!35] fill between [of = B and A];

\addplot[draw=none, name path = C, forget plot] %
                coordinates {
                   (0,0)
                   (14,0)

                   };

            \addplot[draw=none,name path = D, forget plot] %
                coordinates {
                   (0,4)
                   (4,4)
                   (4,5)
                   (7,5)
                   (7,4)
                   (14,4)

                };

\addplot [black!75] fill between [of = C and D];

 \addplot[only marks, mark options={fill=black, mark size=2.5pt},mark=triangle*] %
                coordinates {
                     (4,-1)
                     };

 \addplot[only marks, mark options={mark size=2.5pt},mark=star] %
                coordinates {
                     (7,-1)
                     };

                \addlegendentry{Installed base size}
                \addlegendentry{Installed CM parts}
                 \addlegendentry{Installed AM parts}
                \addlegendentry{CM part failure}
                \addlegendentry{AM part failure}

            \end{axis}
    \end{tikzpicture}
     \caption{Failure process for an exemplary installed base.}\label{fig:exampleprocess}
\end{figure}

Figure~\ref{fig:exampleprocess} depicts an exemplary failure process for an installed base. At time $t=0$, all parts in the installed base of size $10$ are operational. The installed base consist of $4$ AM parts and $6$ CM parts at $t=0$. At $t=4$, one CM part fails, which is directly replaced by an AM part that is on-hand. This replacement changes the installed base composition, and thus the expected number of failures. At $t=7$, an AM part fails. Now, there is no stock on-hand to directly replace the part, so the part is backordered. {Note that we allow the stocking of AM parts, i.e., we allow both make-to-stock and make-to-order of AM. This is necessary because we consider downtime prone assets that have high backorder costs, i.e., in certain instances it is too costly to wait for a couple of hours for an AM part to be printed.}

\subsection{Notation}
\label{subsec:notation}

In this section, we introduce the notation for (i) input, (ii) state, (iii) decisions, (iv) output variables necessary to obtain near-optimal decisions, and (v) auxiliary variables to support the calculation of transitions and to limit the size of the state and decision spaces.

The number of parts in the installed base is equal to $N$, and $S$ denotes the total number of items circulating in the inventory system, {necessary to obtain a tractable model for the exact benchmark}. The number of CM (AM) failures per period has mean $\mu_{C}$ and variance $Var_{C}$ ($\mu_{A}$ and $Var_{A}$). The \gls{CM} (AM) replenishment lead time is denoted by $l_{C}$ ($l_{A}$). For each \gls{CM} (\gls{AM}) order, we encounter procurement costs $c_{C}$ ($c_{A}$) per item. Furthermore, $k_{C}$ ($k_{A}$) denotes the fixed \gls{CM} (\gls{AM}) order costs. Orders for the \gls{CM} supply channel are placed in integer multiples of a given batch size $Q_{C}$. For each failure, the service provider incurs maintenance costs $m$, which includes costs related to transportation of spare parts, engineer time, and materials for diagnosis and repair. These costs must be included in our model, as our sourcing decisions affect future demand and, consequently, future maintenance costs. In standard dual sourcing models, these costs are typically omitted. Unfulfilled demand is backlogged and causes backorder costs $b$ per period. {We consider the costs of downtime per period to be absorbed by the backorder costs.} Furthermore, we use holding costs $h$ per period, independent of the item type. 

We use $t$ to indicate the period. A state $\mathbf{i}_{t} \in \mathcal{I}$ at period $t$ is represented by the following discrete $6$-dimensional integer vector:
\begin{equation}
    \mathbf{i}_t = [n_{C,t}, n_{A,t}, s_{C,t}, s_{A,t} , u_{C,t,t^\prime}, u_{A,t,t^\prime} ],
\end{equation}
where $n_{C,t}$ and $n_{A,t}$ are the number of operating CM and AM parts, respectively. The on-hand stock of CM (AM) parts is denoted by $s_{C,t}$ ($s_{A,t}$), and the number of batches (items) ordered at a previous period $t^\prime$ is recorded by $u_{C,t,t^\prime}$ ($u_{A,t,t^\prime}$). Note that $u_{C,t,t^\prime}$ and $u_{A,t,t^\prime}$ are used to track the pipeline of incoming replenishment orders know at $t$. They denote the pipeline of orders for the \gls{CM} and \gls{AM} parts in the last $l_C-1$ ($l_A-1$) periods, recorded before arrival of the orders from $l_C$ ($l_A$) periods ago and the new order at time $t$. So, $t^\prime \in [t-l_C,t-1]$ for CM and $t^\prime \in [t-l_A,t-1]$ for AM.

The objective it to find a policy $\pi: \mathcal{I} \rightarrow \mathcal{A}$ that minimise average costs per time unit. Here $\mathcal{A}$ is the discrete decision space, which consist of two decisions: the decision variables $x_C$ denote the number of CM batches, and $x_A$ denotes the number of AM items ordered in period $t$. As output of a policy, we denote the long-term average service costs of policy $\pi$ by $g(\pi)$, and the value of a state $\mathbf{i}_{t}$ if we follow policy $\pi$ by $v(\mathbf{i}_{t}, \pi)$. See Section~\ref{subsec:CostFun} for the definition of $v(\mathbf{i}_{t}, \pi)$.

\begin{table}[!t]
\caption{Notation overview}
\label{tab:1}
\resizebox{\textwidth}{!}{%
\begin{tabular}{ll}
\toprule
\textbf{Input variables} & \\
 $N$ & Number of parts in the installed base\\
  $S$ & Maximum number of spare parts circulating in the inventory system \\
 $\mu_{C}$ & Mean number of failures of a \gls{CM} item per period \\
 $\mu_{A}$ & Mean number of failures of an \gls{AM} item per period \\
 $Var_{C}$ & Variance of the number of failures of a \gls{CM} item per period \\ 
 $Var_{A}$ & Variance of the number of failures of an \gls{AM} item per period\\
 $l_{C}$ & Deterministic \gls{CM} replenishment lead time \\
 $l_{A}$ & Deterministic \gls{AM} replenishment lead time \\
 $c_{C}$ &\gls{CM} piece price \\
 $c_{A}$ &\gls{AM} piece price \\
 $k_{C}$ & Fixed \gls{CM} order costs\\
 $k_{A}$ & Fixed \gls{AM} order costs\\
 $Q_{C}$ &  Fixed \gls{CM} batch size\\
 $m$ & Maintenance costs, i.e., costs to replace a failed part\\
 $h$ & Holding cost per item in stock at the beginning of the period\\
 $b$ & Backorder cost per item backlogged at the beginning of a period\\
 \cline{2-2}
  & \\
\textbf{State variables} & \\
$\mathbf{i}_{t}\in \mathcal{I}_{t}$ & System state tuple, $\mathbf{i}_{t}=[n_{C,t}, n_{A,t}, s_{C,t}, s_{A,t}, u_{C,t,t^\prime}, u_{A,t,t^\prime}]$ \\ 
 $n_{C,t}$ & Number of operating \gls{CM} parts at the beginning of period $t$\\
 $n_{A,t}$ & Number of operating \gls{AM} parts at the beginning of period $t$\\
 $s_{C,t}$ &\gls{CM} on-hand stock at the beginning of period $t$\\
 $s_{A,t}$ &\gls{AM} on-hand stock at the beginning of period $t$\\
 $u_{C,t, t^\prime}$ & The number of \gls{CM} batches ordered at time $t^\prime$, known at time $t$, where $t^\prime \in [t-l_C,t-1]$\\
 $u_{A,t,t^\prime}$ & The number of \gls{AM} items ordered at time $t^\prime$, known at time $t$, where $t^\prime \in [t-l_A,t-1]$\\
 \cline{2-2} 
 & \\
\textbf{Decision variables} & \\
 $x_{C}$ & Number of \gls{CM} batches to order at the end of a period $t$ \\
 $x_{A}$ & Number of \gls{AM} parts to order at the end of a period $t$ \\\cline{2-2}
 & \\
 \textbf{Output variables} & \\
  $g(\pi)$ & Long-run average service cost of policy $\pi$\\
  $v(\mathbf{i}_{t}, \pi)$ & Value if we are in state $\mathbf{i}_{t}$ and follow policy $\pi$ \\
   \cline{2-2} 
   & \\
\textbf{Auxiliary variables} & \\
 $B_{t}$ & Number of backorders at the beginning of period $t$\\
 $IP_{t}$ &  Inventory position at the beginning of period $t$\\
 $k_{C,t}$ & Number of realised CM failures in period $t$, having probability density $p_{C}(k_{C,t}|n_{C,t})$\\
  $k_{A,t}$ & Number of realised AM failures in period $t$, having probability density $p_{A}(k_{A,t}|n_{A,t})$\\
 
 \bottomrule
\end{tabular}}
\end{table}

To aid in the calculations of the transitions from state $\mathbf{i}_{t}$ to $\mathbf{i}_{t+1}$ and to limit the sizes of the state and decision spaces, we define several auxiliary variables. These include the number of backorders at the beginning of period $t$, denoted by $B_t$, the inventory position at the beginning of period $t$, denoted by $IP_t$, and the realised number of \gls{CM} and \gls{AM} failures in a period, denoted by $k_{C,t}$ and $k_{A,t}$, respectively. Table~\ref{tab:1} summarises the notation.

\subsection{State Space Constraints}
\label{subsec:State space}

We calculate the backorders ($B_t$) and inventory positon ($IP_t$) at the start of period $t$. We express the number of backorders at the start of period $t$ by Equation~\ref{eq:backorder} and the inventory position at the start of period $t$ by Equation~\ref{eq:ip}. 
\begin{align}
B_{t}&=N-n_{C,t}-n_{A,t}, \label{eq:backorder}\\
IP_t &= s_{C,t} + s_{A,t} + \sum_{t^\prime=t-l_C}^{t^\prime=t-1}u_{C,t,t^\prime}Q_C + \sum_{t^\prime=t-l_A}^{t^\prime=t-1} u_{a,t,t^\prime} - B_t,\label{eq:ip}
\end{align}
where backorders are calculated by subtracting the items in operation from the installed base size, and the inventory position is the sum of on-hand stock and orders in the pipeline minus the backorders. We next constrain the state space in period $t$ by: 
\begin{subequations}
\begin{align}
\mathcal{I} =\{(&n_{C,t}, n_{A,t}, s_{C,t}, s_{A,t}, u_{C,t,t^\prime}, u_{A,t,t^\prime}):\nonumber
\\
&0\leq n_{C,t} + n_{A,t}\leq N, \label{eq1.1}
\\
&IP_{t} \leq S \label{eq1.2}
\}.
\end{align}
\end{subequations}%

Constraint~\ref{eq1.1} ensures that the number of installed items cannot become negative nor exceed the installed base size. Constraint~\ref{eq1.2} limits the number of items in the inventory system. That is, the inventory position at any period $t$ may not exceed some predefined value $S$. This constraint is necessary to obtain a finite state space. {For the exact policy, we determine $S$ numerically by finding the value for $S$ where the probability of having more than $S$ in the replenishment pipeline is only marginal.} For all other policies, $S$ is set to a large number such that the decision freedom of the policy is not too restricted. We refer to Appendix~\ref{appendix:hyper} for more details.

\subsection{Decision Space Constraints}
\label{subsec:decisio}
The decision is defined by $x_{C}$ and $x_{A}$, where $x_{C}$ describes the number of \gls{CM} batches and $x_{A}$ the number of \gls{AM} parts that are ordered. Using state information and the relationship between both decision variables, we limit the decision space to:
\begin{align}
\mathcal{A} =\{(&x_{C}, x_{A}):\nonumber
\\
&0\leq x_{C}Q_c+x_A\leq S-IP_{t} \label{eq2.3}
\}.
\end{align}

Constraint~\ref{eq2.3} ensures that the total order quantity cannot be negative nor raise the inventory position above the maximum number of spare parts in the system $S$.

\subsection{Transitions}
\label{subsec:transition}
We first describe the transition function for each state variable, followed by the computation of transition probabilities. Recall that the realised number of \gls{CM} and \gls{AM} failures in a period are denoted by $k_{C,t}$ and $k_{A,t}$, respectively. The number of CM (AM) items in the installed base at the beginning of period $t+1$ is equal to the number at the beginning of period $t$, minus the number of CM (AM) parts that failed, plus the number of CM (AM) parts that replaced the failed parts, and any backorders. The last component is complex, as it depends on the availability of parts (AM and CM) and on which part has the lowest failure rate, see Assumption~5. The installed base composition at the beginning of period $t+1$ is then obtained as follows:
\begin{eqnarray}
n_{C,t+1}= n_{C,t}-k_{C,t}+y_{C,t}+z_{C,t},\label{eq6}\\ 
n_{A,t+1}= n_{A,t}-k_{A,t}+y_{A,t}+z_{A,t},
\end{eqnarray}
where the helper variables $y_{C,t}$ ($y_{A,t}$) denote the maximum number of \gls{CM}(AM) items that are available \textit{before} a possible replenishment arrives at the end of period $t$, and $z_{C,t}$ ($z_{A,t}$) the maximum number of \gls{CM}(AM) items that are available \textit{after} a possible order arrival. We need this distinction, since we assume that items with lower failure rate are consumed first with on-hand stock available before the arrival of replenishments; cf. Assumption~5. In Appendix~\ref{app:transitions} we detail how we calculate the values of $y_{C,t}$, $y_{A,t}$, $z_{C,t}$, and $z_{A,t}$. For the on-hand stock in period $t+1$, we have:
\begin{eqnarray}
s_{C,t+1}= s_{C,t} + A_{C,t} - y_{C,t} - z_{C,t},\\ 
s_{A,t+1}= s_{A,t} + A_{A,t} - y_{A,t} - z_{A,t},
\end{eqnarray}
\noindent where $A_{C,t}$ ($A_{A,t}$) equal the number of CM (AM) replenishment items that may arrive at the end of period $t$, see Appendix~\ref{app:transitions} for its definition. The order quantities are updated by recording all orders still to arrive:
\begin{eqnarray}
u_{C,t+1,t^\prime} =  x_{C,t^\prime} \quad \forall t^\prime \in [t-l_C+1,t],  \\
u_{A,t+1,t^\prime} = x_{A,t^\prime} \quad \forall t^\prime \in [t-l_A+1,t],\label{eq11}
\end{eqnarray}
$u_{C,t+1,t^\prime}$ and $u_{A,t+1,t^\prime}$ may contain zeros, indicating that no order was placed at $t^\prime$.

Next, we derive the transition probabilities of the items in operation. Equations~\ref{eq6}-\ref{eq11} show the deterministic transitions dependent on the decisions. The only stochastic component of the transitions are failures ($k_{C,t}$ and $k_{A,t}$) in a review period, which influence the parts in operation ($n_{C,t}$ and $n_{A,t}$). Using Assumption 3 from Section~\ref{sec:prob}, the probability mass functions $\mathbb{P}_{C}(k_{C,t}|n_{C,t})$ and $\mathbb{P}_{A}(k_{A,t}|n_{A,t})$ are characterised by $\mathbb{E}[k_{x}]=\mu_{x}n_{x,t}$ and $Var[k_{x}]=V_{x}n_{x,t}$ for production method $x \in \{C,A\}$. Assumption~3 may result in $k_{C,t}>n_{C,t}$ and $k_{A,t}>n_{A,t}$ with low probability, as we do not track the installed base during a review period. We address this by assigning the probability masses to cases where $k_{C,t}=n_{C,t}$ and $k_{A,t}=n_{A,t}$, respectively, thus limiting failures to $n_{C,t}$ and $n_{A,t}$. This simplification is reasonable since the probability of failures exceeding the installed base size within one review period is negligible for realistic instances. We define the transition probabilities using $k_{C,t}$ and $k_{A,t}$:
\begin{equation}
\mathbb{P}\left(n_{C,t+1},n_{A,t+1}|n_{C,t},n_{A,t}\right)=
\mathbb{P}_{C}\left(k_{C,t}|n_{C,t})\mathbb{P}_{A}(k_{A,t}|n_{A,t}\right).
\end{equation}

\subsection{Cost Function}
\label{subsec:CostFun}
We minimise the average service costs per period, $C(\mathbf{i}_{t},x_C, x_A)$, consisting of purchasing costs $P(x_C, x_A)$ (fixed and variable), holding costs $H(\mathbf{i}_{t})$, backorder costs $BO(\mathbf{i}_{t})$ and maintenance costs $M(\mathbf{i}_{t})$ over an infinite horizon. 
The purchasing costs are equal to:
\begin{equation}
P(x_C, x_A) = \mathbf{1}_{x_{C}>0}k_C+ \mathbf{1}_{x_{A}>0}k_A + c_{C}x_{C}Q_C+c_{A}x_{A},
\end{equation}
where $\mathbf{1}_{x_{C}>0}$ ($\mathbf{1}_{x_{A}>0}$) is the indicator function, which is equal to 1 if $x_{C}>0$ ($x_{A}>0$) and 0 otherwise, and used to determine the fixed order costs.
The holding costs and the expected backorder costs are equal to:
\begin{align}
H(\mathbf{i}_{t}) &= h(s_{C,t}+s_{A,t}),\\
BO(\mathbf{i}_{t}) &= \max(k_{C,t}+k_{A,t}+B_{t}-s_{C,t}-s_{A,t},0)b.
\end{align}

We encounter expected maintenance cost for each failure and assign these costs to the period in which the failures occurs, and not to the period where the repair is made, thus we have:
\begin{equation}
M(\mathbf{i}_{t}) = m(\mu_{C}n_{C,t}+\mu_{A}n_{A,t}).  
\end{equation}

If the policy $\pi$ is optimal, the optimal long-run average service costs $g(\pi)$ and the value function $v(\mathbf{i}_{t})$ satisfy the Bellman equation:
\begin{equation}
\label{eq:bellEQ}
v(\mathbf{i}_{t})=\min_{(x_C, x_A)\in\mathcal{A}}\{C(\mathbf{i}_{t},x_C, x_A)-g(\pi)+\sum_{\mathbf{i}_{t+1}\in\mathcal{I}_{t+1}} \mathbb{P}(\mathbf{i}_{t+1}|\mathbf{i}_{t}, x_C, x_A)V(\mathbf{i}_{t+1})\}, \quad \forall \mathbf{i}_{t}\in\mathcal{I},
\end{equation}
where $v(\mathbf{i}_{t+1})$ denotes the downstream (long-term) value of the state $\mathbf{i}_{t+1}$ if we follow policy $\pi$.

\section{Solution Approaches}\label{sec:Approach}
The size of the state space makes exact analysis impractical for medium to large problem sizes, allowing benchmarking against an exact policy only for small instances. In Appendix~\ref{app:simpModel}, we outline the exact solution procedure. Section~\ref{subsec:Iter} explains a heuristic iterative method that allows for extending a standard discrete-time dual sourcing heuristic such that is can deal with endogenous demand. Section~\ref{sct:rl} outlines the three studied RL algorithms. Finally, Section~\ref{subsect:EPL} presents a novel approach, denoted by \gls{EPL}, to enhance RL methods.

\subsection{\gls{IWA}}\label{subsec:Iter}

We propose an algorithm to obtain a policy that balances \gls{CM} and \gls{AM} sourcing. The key idea is as follows. We use a standard dual sourcing policy $\pi^S$ with a single demand rate, based on a fixed fraction of \gls{AM} and \gls{CM} parts in the installed base. This demand rate is computed as a weighted average of the \gls{CM} and \gls{AM} failure rates, with weights iteratively adjusted.

In each iteration, the algorithm outputs the fraction of \gls{AM} parts from the total parts sourced using \gls{AM} and \gls{CM}, (AM/(AM+CM)). This output gives us a new single weighted average failure rate, which is then fed back into the model to estimate the fraction of \gls{AM} and \gls{CM} parts in the installed base. This fraction, denoted by $\gamma_{j}$ (where $j$ is the iteration index of the \gls{IWA} process), offers an updated estimate for spare parts demand, differing from our initial hypothesis, denoted by $\rho_j$.

\begin{figure}[hbtp]
    \centering
    \resizebox{\textwidth}{!}{%
    \begin{tikzpicture}[scale=0.8,node distance=5cm, auto]
    % Define style for boxes
    \tikzstyle{block} = [rectangle, draw, text width=10em, text centered, rounded corners, minimum height=4em]

    % Define style for arrows
    \tikzstyle{arrow} = [thick,->,>=stealth]

    % Place nodes
    \node [text width=10em, label={[text width=5em, align=center, yshift=0.0cm, xshift=-1.5cm]above:Start $\gamma_1$}] (block0) at (-2,0) {};
    \node [block] (block1) {Calculate single demand rate parameters $\mathbb{E}[X]_j$ and Var[X]$_j$};
    \node [block, right of=block1] (block2) {Measure $\rho_j$ using $\pi^S$($\mathbb{E}[X]_j$;Var[x]$_j$) };
    \node [block, right of=block2] (block3) {Calculate CM/AM weight $\gamma_{j+1}(\rho_j)$};
     \node  (block4) at (6.2,-2.5) {If not converged, move to next iteration $j=j+1$};

    % Draw edges
     \draw [arrow] (block0) -- (block1);
    \draw [arrow] (block1) -- (block2);
    \draw [arrow] (block2) -- (block3);
   \draw [arrow] (block3) |- ++(0,-2) -| (block1);
\end{tikzpicture}}
    \caption{High-level overview of the \gls{IWA} algorithm.}
    \label{fig:IWA}
\end{figure}
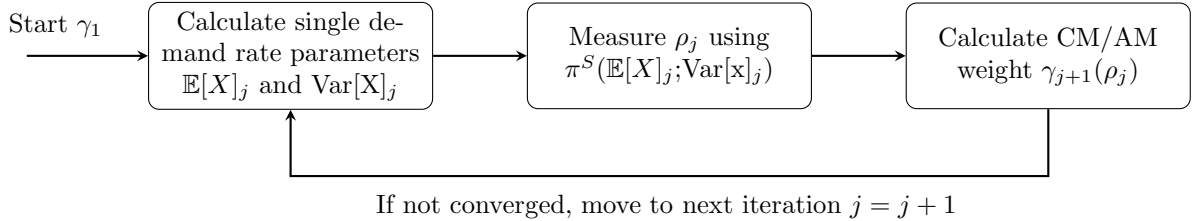

We continue this iterative process until convergence, i.e., until the fraction of parts sourced by \gls{AM} and \gls{CM} no longer changes. A high-level overview of this iterative process is depicted in Figure~\ref{fig:IWA}.

Next, we provide more details about \gls{IWA}. Suppose we have an estimate for the average fraction $\gamma_{j}$ of \gls{CM} and \gls{AM} items in operation during iteration $j$. We can then compute the expectation $\mathbb{E}[X]$ and the variance $\mathrm{Var}[X]$ of a random variable $X$ that measures the number of failures per period of an arbitrary item in the installed base. By applying standard conditioning rules for the mean and variance, we obtain:
\begin{align}
    \mathbb{E}[X] &= \gamma_{j}\mu_{A} + (1-\gamma_{j})\mu_{C},\label{6eq:expect}\\
    \mathrm{Var}[X] &= \gamma_{j} V_{A} + (1-\gamma_{j}) V_{C} + \gamma_{j}(1-\gamma_{j})(\mu_{A}-\mu_{C})^{2}.\label{6eq:var}
\end{align}

Next, to find the new estimate of the average fraction of \gls{CM} and \gls{AM} items in operation, $\gamma_{j+1}$, for iteration $j+1$, we measure the fraction $\rho_{j}$ of sourced \gls{AM} items using a specified dual sourcing solution procedure, and the previously computed $\mathbb{E}[X]$ and $\mathrm{Var}[X]$. Note that $\rho_{j}$ must equal the fraction of \gls{AM} items entering or leaving the installed base, as shown in Equation~\ref{6eq:Roh}. Finally, we solve Equation~\ref{6eq:Roh} for $\gamma_{j+1}$ and obtain Equation~\ref{6eq:Gamma}:
\begin{align}
    \rho_{j} &= \dfrac{\gamma_{j+1} \mu_{A}}{\gamma_{j+1} \mu_{A} + (1-\gamma_{j+1}) \mu_{C}},\label{6eq:Roh}\\
    \gamma_{j+1} &= \dfrac{\rho_{j} \mu_{C}}{(1-\rho_{j}) \mu_{A} + \rho_{j} \mu_{C}}.\label{6eq:Gamma}    
\end{align}

This iterative procedure continues until the difference between two subsequent values of $\gamma$ is less than a specified threshold. This stopping criterium is needed to prevent cycling between policies, caused by discretisation effects. The complete algorithm outline is provided in Appendix~\ref{app:ira}. We detail the policy $\pi^S$ in Section~\ref{sec:Experiments}.

\subsection{\gls{RL} approach}\label{sct:rl}

\gls{RL} is a framework for modelling sequential decision-making problems, where an agent learns to minimise costs by interacting with an environment. The agent observes the state, makes decisions, and receives feedback in the form of rewards. We implement three RL algorithms from the main RL classes: policy-based (\gls{DCL}, \citet{temizoz2023deep}), value-based (\gls{AVI}, \citet{powellBookNew}), and actor-critic (\gls{PPO}, \citet{schulman2017proximal}). \gls{DCL} and \gls{PPO} use neural networks, while \gls{AVI} uses linear regression. Implementation details for DCL, AVI, and PPO are given in Appendix~\ref{subsec:DCL},~E, and~G, respectively.

For sake of conciseness, we provide generic preliminaries for RL. A policy \(\pi\) in \gls{RL} determines decisions based on the current state \(\mathbf{i}_t\). Policies can be deterministic (DCL and AVI) or stochastic (PPO). The agent's objective is to learn an optimal policy \(\pi^*\) that maximises expected cumulative rewards. The RL training process involves: 
\begin{enumerate}
    \item \textbf{Initialisation}: Initialise the policy \(\pi\) and, if applicable (AVI and PPO), the value function \(v\).
    \item \textbf{Interaction with Environment}: Simulate decisions based on the policy.
    \item \textbf{Observation and Rewards}: Observe resulting states and obtain rewards.
    \item \textbf{Policy Update}: Update the policy (DCL and PPO) and value function (PPO and AVI) based on observed transitions and rewards.
\end{enumerate}
Value-based methods estimate value functions \(v(\cdot)\) to derive optimal policies, e.g., \gls{AVI}. Policy-based methods, like DCL, directly optimise the policy \(\pi(x_t|\mathbf{i}_t)\) to maximise expected rewards. Actor-critic methods, such as PPO, combine value-based and policy-based approaches, using an actor as the policy and a critic to evaluate decisions. For more details on RL, refer to \citet{sutton} and \citet{powellBookNew}.
RL has the advantage that it scales to larger problem sizes with large installed base sizes. However, as reviewed in Section~\ref{litrev:RL}, RL policies are known to be non-robust and therefore, a single trained policy cannot be easily used for multiple SKUs. This means that a policy has to be trained for every SKU separately, while a company may run thousands of SKUs that are technically suitable for AM. In the next section, we present a simple approach that tackles this problem.

\subsection{\gls{EPL}}\label{subsect:EPL}

We present an approach called \gls{EPL}, which significantly reduces the overall training burden by training a single policy applicable to multiple SKUs. The basic idea is to train an RL policy using a fine grid of SKU parameter values, and sampling from this grid during training. We assume that similar parameter settings yield comparable optimal decisions, enabling more efficient joint training. As \emph{endogenous} problem parameters change during training, they become part of the state observation.

We consider two primary strategies for training the RL policy embedded in EPL. First, we can train a single policy on a given set of SKUs and then use this global learning model specifically for those SKUs. Second, we can train (cross-learn) a policy on a grid of parameter settings, allowing it to be applied to any settings within this grid, even those not explicitly trained on. This strategy aims to create a more generalised policy capable of handling variations within the parameter space, e.g., the application to new products with similar characteristics.

Choosing the second strategy requires careful consideration of grid design and parameter settings selection. The grid must be detailed enough to capture variability without becoming computationally infeasible. While this initial grid design and training method aim to demonstrate the potential of our \gls{EPL} approach, optimising the selection of training instances and including strategies for refining the grid, could be explored further in subsequent studies (as discussed in Section~\ref{sec:Conclusion}). Below, we provide more details of our \gls{EPL} approach.

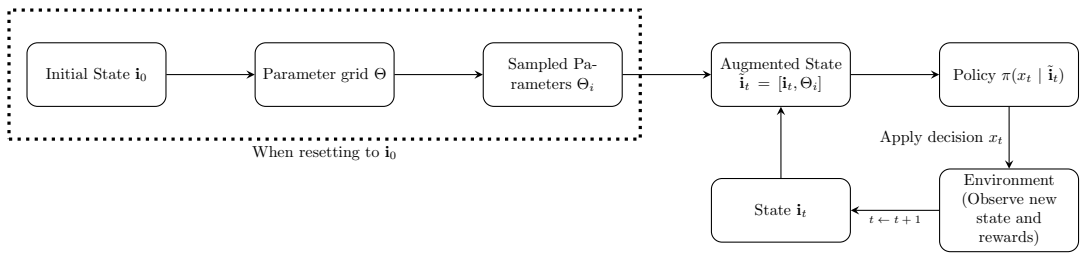
\begin{figure}[h]
    \centering
    \begin{tikzpicture}[scale=0.6, every node/.style={scale=0.6}]
        % Styles
        \tikzstyle{block} = [rectangle, draw, text width=8em, text centered, rounded corners, minimum height=4em]
        \tikzstyle{line} = [draw, -stealth]

        % Nodes
        \node [block] (initial) {Initial State \(\mathbf{i}_0\)};
        \node [block, right of=initial, node distance=5cm] (sampling) {Parameter grid \(\Theta\)};
        \node [block, right of=sampling, node distance=5cm] (params) {Sampled Parameters \(\Theta_i\)};
        \node [block, right of=params, node distance=5cm] (augmented) {Augmented State \(\tilde{\mathbf{i}}_t = [\mathbf{i}_t, \Theta_i]\)};
        \node [block, below of=augmented, node distance=3cm] (state) {State \(\mathbf{i}_t\)};
        \node [block, right of=augmented, node distance=5cm] (policy) {Policy \(\pi(x_t \mid \tilde{\mathbf{i}}_t)\)};
        \node [block, below of=policy, node distance=3cm] (env) {Environment\\(Observe new state and rewards)};

        % Fit box around specific nodes
        \node[fit=(initial) (sampling) (params), draw, very thick, dotted, inner ysep=1cm,inner xsep=3cm, label=below:{When resetting to \(\mathbf{i}_0\)}] {};

        % Arrows
        \path [line] (initial) -- (sampling);
        \path [line] (sampling) -- (params);
        \path [line] (params) -- (augmented);
        \path [line] (state) -- (augmented);
        \path [line] (augmented) -- (policy);
        \path [line] (policy) -- (env) node[midway, left] {Apply decision $x_t$};
        \path [line] (env) -- (state)  node[midway, below] {\scriptsize{\( t \leftarrow t+1 \)}};
    \end{tikzpicture}
    \caption{Overview of a general RL training procedure embedded in EPL.}
    \label{fig:epl_approach}
\end{figure}

We denote the input parameters by $\theta \in \Theta$, where $\Theta$ represents the set of all parameters, e.g., installed base size, backorder costs, etc. Each parameter $\theta_j$ ($j = 1, 2, \ldots, n$) has a set of possible values $\mathcal{V}_j = \{v_{1}^j, v_{2}^j, \ldots, v_{m}^j\}$, derived from the known information about all SKUs. In each iteration $i$, a value is sampled for each parameter, denoted as $\theta_{ji}$. Thus, the set of sampled values in iteration $i$ is:
\[
    \Theta_i = (\theta_{1i}, \theta_{2i}, \ldots, \theta_{ni}),
\]
where $\theta_{ji} \in \mathcal{V}_j$. By treating problem parameters as part of the state space, we allow the learning policy to account for variations in problem settings, improving generalisation and robustness. The methodology can be summarised as:
\begin{enumerate}
    \item \textbf{Problem parameter sampling}: 
    Define a fine grid over the dual sourcing problem parameters and sample parameter values \(\theta_{ji}\) from the set \(\mathcal{V}_j\) during training.
    \item \textbf{State space augmentation}:
    Embed the sampled problem parameters into the state space. This augmentation \(\tilde{\mathbf{i}}_t = [\mathbf{i}_t, \Theta_i]\) provides the learning agent with contextual information necessary for making informed decisions.    
    \item \textbf{Policy training}:
    Train the learning agent using a standard learning algorithm with the augmented state space, optimising decisions by considering both the immediate state and the underlying problem parameters.
    \begin{equation}
         \pi(x_t \mid \tilde{\mathbf{i}}_t) = \pi(x_t \mid [\mathbf{i}_t, \Theta_i]).
    \end{equation}
\end{enumerate}
To ensure stability during training, new problem parameters are only sampled when resetting to an initial state, keeping $\Theta_i$ fixed during a decision-making horizon.
Figure~\ref{fig:epl_approach} provides a visual representation of a typical \gls{RL} training loop, including the sequential decision-making process, embedded in EPL. In Section~\ref{sect:paramEPL} we explain how we define $\Theta$ specifically for our study.

% \medbreak
% \begin{itemize}
% \item $IP_{C,t}$ and $IP_{A,t}$: the inventory positions of \gls{CM} and \gls{AM} parts as defined by Equation~\ref{eq:invenP_CM} and Equation~\ref{eq:invenP_AM}, respectively.
% \item $n_{C,t}$ and $n_{A,t}$: The number of \gls{CM} and \gls{AM} operating parts, repsectively.
% \item $s_{C,t}$ and $s_{A,t}$: the \gls{CM} and \gls{AM} on-hand stock, respectively.
% \end{itemize} 
% \medbreak
% We do not include past order information, as this information is inherently contained in the inventory positions of \gls{CM} and \gls{AM} parts. In contrast to ADP, we calculate features over the ``normal'' state, i.e., the pre-decision state.

Problem parameters are sampled uniformly from a defined grid each time we reset to an initial state. To create the grid:
\begin{enumerate}
    \item Collect SKU data and observe all parameter values.
    \item Calculate key statistics (min, max, and percentiles) for each parameter.
    \item Define grid points by combining the min, max, and selected percentiles (with the number of percentiles as an adjustable parameter).
\end{enumerate}
We define a new hyperparameter $\kappa_j$ for the number of percentiles for each parameter $j$. Increasing $\kappa_j$ makes the grid finer and potentially improves the policy but requires more training effort. The values for $\kappa_j$ in our experiments are detailed in Appendix~\ref{appendix:hyper}. 

During policy evaluation, we do not sample parameters but evaluate specific settings, ensuring that the minimum and maximum sampled values during training cover the problem parameters' range. This ensures that the learned policy interpolates within known bounds rather than extrapolating beyond them. Next, we discuss the changes needed in our studied RL algorithms to support the embedding in EPL.

\subsubsection{Adaptations for EPL}\label{sect:paramEPL}

Two main adaptations are needed to support embedding in EPL compared to standard RL approaches: 
\begin{enumerate}
    \item \textbf{Complexity of Q-value estimation}: Adding more state variables makes the Q-value distribution more complex, requiring a more powerful estimator. This is achieved by increasing the neural network size (if applicable) by adding more hidden layers.
    \item \textbf{Increased training requirements}: A larger state space requires more samples and longer training runs, for details, see Appendix~\ref{appendix:hyper}.
\end{enumerate}
Despite these challenges, EPL's advantage of training a single policy for multiple SKUs significantly reduces the overall training burden compared to separate policies for each SKU. If successful, this approach eliminates the need to retrain for new SKUs with parameter settings within the grid. We apply EPL for the following input parameters: $N$, $\mu_{C}$, $\mu_{A}$, $Var_{C}$, $Var_{A}$, $l_{C}$, $l_{A}$, $c_{C}$, $c_{A}$, $k_{C}$, $k_{A}$, $h$, $b$, and $m$.

\section{Numerical Experiments}\label{sec:Experiments}
In this section, we study the performance of \gls{IWA} and the RL approaches. To summarise, we compare the following policies:
\begin{itemize}
    \item \gls{BSP}: a single source base stock baseline that orders \gls{AM} or \gls{CM}, whichever is cheaper.
    \item \gls{IWA}: our proposed heuristic.
    \item \gls{AVI}: using linear regression as value function, as proposed in, e.g., \citet{powellBookNew}.
    \item \gls{DCL}: using fully connected neural networks as policy, see \citet{temizoz2023deep}.
    \item \gls{PPO}: with fully connected neural networks as actor and critic, see \citet{schulman2017proximal}
\end{itemize}
For IWA, we need to define a standard dual sourcing method to be used as supporting policy $\pi^S$. For the smaller tractable instances (Section~\ref{sec:synth}), we employ an exact policy $\pi^s$, see Appendix~\ref{app:simpModel} for more details. For the large instance (Section~\ref{sec:energy}), we use an approximate dual index policy $\pi^S$, as proposed by \citet{Veeraraghavan2008NowSystems}. For more details, we refer to Appendix~\ref{app:ira}. The complete explanation {and implementation details} of \gls{AVI}, \gls{DCL}, and \gls{PPO} can be found in Appendix~\ref{subsec:ADP},~\ref{subsec:DCL}, and~\ref{app:PPO}, respectively. 

Additionally, we test all learning frameworks embedded in our proposed \gls{EPL} approach, we denote these policies as AVI$^\textrm{EPL}$, \gls{DCL}$^\textrm{EPL}$, and \gls{PPO}$^\textrm{EPL}$. We employ two primary training strategies: training on a specific set of SKUs (Section~\ref{sec:synth}) and training on a grid of parameter settings (Section~\ref{sec:energy}). All problems and algorithms are implemented in C\texttt{++}, the neural network models are implemented in LibTorch, a C\texttt{++} library version of PyTorch \citep{pytorch}. The \gls{PPO} algorithm is implemented using PyBindings with the Tianshou RL library \citep{tianshou}. Computations are conducted on a single thin CPU node. The node is equipped with a $2.6$ GHz AMD Genoa 7H12 processor, has $192$ CPU cores, and $384$ GB of memory. All hyperparameter settings are reported in Appendix~\ref{appendix:hyper}. We only tune hyperparameters once on a single representative instance and do not tune afterwards. %In Appendix~\ref{app:convergence} we provide convergence curves for the learned methods.

In Section~\ref{sec:synth}, we provide the results {for a case of $10$ instances} using synthetic data, and in Section~\ref{sec:energy}, we provide the results for the real-world case in the energy sector, {which consist of $1215$ large instances}. The synthetic case is intended to study more fundamental problem characteristics and allows us to benchmark against an exact policy. The energy case shows the performance of our policies in a real-world setting, which allows us to distill managerial insights.

\subsection{Synthetic Case}
\label{sec:synth}
In this section, we discuss the experimental design of our case study based on synthetic data in Section~\ref{sect:synthexp} and provide results in Section~\ref{subsec:performsynth}. Our experiments aim to determine which approach works best and under what conditions.

\subsubsection{Synthetic Case Experimental Design}\label{sect:synthexp}

In our experiments, we first study small synthetic instances with an installed base size of $N=7$ slow moving items. For these small instances, computing the optimal policy is feasible using a standard policy iteration algorithm, see Appendix~\ref{app:simpModel}. The instance parameters are summarised in Table~\ref{tab:synthScen}. We consider two primary scenarios for the synthetic instances:
\begin{enumerate}
    \item  \textbf{Scenario 1}: The CM item is expensive, fails rarely, and has a long lead time.
    \item \textbf{Scenario 2}: The CM item is cheaper, fails more often, and has shorter lead times.
\end{enumerate}
In the second scenario, we set the AM parameter values to higher purchase costs, higher failure rates, but significantly shorter lead times compared to the first scenario. We do not consider fixed order costs for AM ($k_A=0$). For the remaining parameters, we evaluate settings with high and low maintenance, holding, and backorder costs. Backorder costs are determined using the fill rate ($b\slash(b+h)$). Demand per item follows a logarithmic compound Poisson process, ensuring the variance-to-mean ratio is $\geq 1$.
\begin{table}[hbtp]
\caption{Experimental design scenarios}\label{tab:synthScen}
\centering
\begin{tabular}{lll}
\toprule
\textbf{Parameter} & \textbf{Scenario 1} & \textbf{Scenario 2} \\ \midrule
\textbf{CM Item} &  &  \\
Piece price ($c_C$) & 5000 euros & 1000 euros \\ 
Order cost ($k_{C}$) & 2000 euros & 750 euros \\ 
Failure rate ($\mu_C$) & 0.01/week (Variance: $2\mu_C$) & 0.025/week (Variance: $2\mu_C$) \\ 
Lead time ($l_C$) & 8 weeks & 4 weeks \\ 
Order quantity ($Q_C$) & 5 & 7 \\ \midrule
\textbf{AM Item} &  &  \\
Piece price ($c_A$) & $2c_C$ & $1.5c_C$ \\ 
Failure rate ($\mu_A$) & $2\mu_C$ (Variance: $2\mu_A$) & $1.5\mu_C$ (Variance: $3\mu_A$) \\ 
Lead time ($l_A$) & $l_C\slash 4$ & $l_C\slash 2$ \\ \midrule
\textbf{Other Parameters} &  &  \\
Maintenance cost ($m$) & $c_C$ euros & $0.25c_C$ euros \\ 
Holding cost ($h$) & $0.3c_C$ euros/year & $0.2c_C$ euros/year \\ 
Fill rate & 99.5\% & 99\% \\ 
\bottomrule
\end{tabular}
\end{table}

Our setup leads to eight instances; the entire set of instances is summarised in Table~\ref{tab:instance input}. Additionally, we consider two extreme cases (Instances 9 and 10) to evaluate specific conditions: Instance 9 is designed such that it leads to a relatively large usage of AM as emergency source, by increasing the lead time of CM and reducing the AM lead time. Instance 10 has a similar lead time setting, but now the AM parts have a significantly higher failure rate compared to CM parts.

\begin{table}[hbtp]
\centering
\footnotesize
\caption{Instance settings.}
\label{tab:instance input}
\resizebox{\textwidth}{!}{
\begin{tabular}{@{}lllllllllll@{}}
\toprule
 & \multicolumn{10}{c}{\textbf{Instance}} \\ \cmidrule(l){2-11} 
\textbf{Parameter} & 1 & 2 & 3 & 4 & 5 & 6 & 7 & 8 & 9 & 10 \\ \cmidrule(l){2-11} 
$N$ & 7 & 7 & 7 & 7 & 7 & 7 & 7 & 7 & 7 & 7 \\%& 20 & 20 \\
$c_{C}$ & 5000 & 5000 & 5000 & 5000 & 1000 & 1000 & 1000 & 1000 & 2000 & 2000 \\
$k_{C}$ & 2000 & 2000 & 2000 & 2000 & 750 & 750 & 750 & 750 & 2000 & 1000 \\
$\mu_{C}$ & 0.01 & 0.01 & 0.01 & 0.01 & 0.025 & 0.025 & 0.025 & 0.025 & 0.01 & 0.025 \\
$Var_{C}$ & 0.02 & 0.02 & 0.02 & 0.02 & 0.05 & 0.05 & 0.05 & 0.05 & 0.02 & 0.05 \\
$l_{C}$ & 8 & 8 & 8 & 8 & 4 & 4 & 4 & 4 & 10 & 6 \\
$c_A$ & 10000 & 10000 & 7500 & 7500 & 2000 & 2000 & 1500 & 1500 & 3000 & 2400 \\
$\mu_{A}$ & 0.02 & 0.02 & 0.015 & 0.015 & 0.05 & 0.05 & 0.0375 & 0.0375 & 0.015 & 0.125 \\
$Var_{A}$ & 0.04 & 0.04 & 0.045 & 0.045 & 0.1 & 0.1 & 0.1125 & 0.1125 & 0.03 & 0.375 \\
$l_{A}$ & 2 & 2 & 4 & 4 & 1 & 1 & 2 & 2 & 1 & 1 \\
m & 5000 & 1250 & 5000 & 1250 & 1000 & 250 & 1000 & 250 & 500 & 2000 \\
h & 29 & 19 & 29 & 19 & 6 & 4 & 6 & 4 & 12 & 12 \\
b & 5725 & 1899 & 5725 & 1899 & 1145 & 380 & 1145 & 380 & 2290 & 2290 \\
\bottomrule
\end{tabular}}
\end{table}

We report optimality gaps for the ten stylised instances. The optimality gaps are calculated as: $\frac{v_\pi - v^*}{v^*} \cdot 100\%$, where $v_\pi$ is the average cost per period of policy $\pi$, and $v^*$ is the average cost per period for the optimal policy.

\subsubsection{Results for the Synthetic Case}
\label{subsec:performsynth}

{As we stressed in Section~\ref{subsec:Iter}, \gls{IWA} may be applied in combination with most dual sourcing solution procedures since it only relies on an estimate of a weight of items ordered from either supply mode. In this section we apply IWA in combination with an exact policy, see Appendix~\ref{app:simpModel}.
\medbreak
Table~\ref{tab:results} shows the results for each problem instance. We observe that DCL and DCL$^\textrm{EPL}$ are the best performing policies, with an average optimality gap of $0.4\%$. In two instances (Instance 6 and 7), our proposed IWA approach is the single best performing policy. These instances have the smallest gap between the single sourcing BSP policy and the optimal dual sourcing method. This suggests that IWA outperforms other policies when the advantage of dual sourcing over single sourcing is relatively small. However, in Instance 8, where the BSP gap is also relatively small, DCL finds a policy that matches IWA. We conjecture this is due to the lower AM purchase, maintenance, and backorder costs in Instance 8 compared to Instances 6 and 7. 
\begin{observation}
    Generally DCL performs best, and in instances where other algorithms do better, the gains from dual sourcing are minimal. 
\end{observation}
For the remaining 7 instances, either DCL, DCL$^\textrm{EPL}$, or both policies achieve the best performance. AVI, using a linear regression value function, and PPO, using neural networks, are unable to find a performant policy. We find that both policies hardly order AM parts. Due to the low demand rates, the optimal decision for most states is to order nothing. Therefore, the deviating reorder decision of only a few states determines the solution quality. For example, for Instance 4, we find that the optimal decision is to order nothing in more than 93\% of all states. Both AVI and PPO struggle to learn a policy for such sparse reward states, which explains the large gaps of \gls{AVI} and \gls{PPO}. Linear regression (used for AVI) seems not powerful enough to approximate the complex state-decision function and \gls{PPO} is unable to find a competitive policy due to its sampling and updating structure which does not evaluate enough exogenous scenarios, i.e., it is unable to deal with the sparse reward structure.

\begin{table}[t]
    \caption{Optimality gaps of the policies, the best performing policies are \textbf{highlighted}.}
    \label{tab:results}
    \centering
    \resizebox{\textwidth}{!}{
    \begin{tabular}{l c c c c c c c c c }
    \toprule
    Instance & \gls{BSP}& \gls{IWA}& \gls{AVI} & AVI$^\textrm{EPL}$ & \gls{DCL} & \gls{DCL}$^\textrm{EPL}$ & \gls{PPO} & \gls{PPO}$^\textrm{EPL}$ \\
    \midrule 
    1 & 5.9\%  & 0.4\% & 6.2\%  & 5.8\%  & \textbf{0.3\% } &  \textbf{0.3\% }  & 6.4\%  & 5.9\%  \\
    2 & 3.7\%  & 2.4\%  & 3.9\%  & 3.8\%  &\textbf{0.3\% } & \textbf{0.3\% }  & 3.7\%   & 3.7\%  \\
    3 & 13.5\%  & 7.4\%  & 14.1\%  & 13.6\%   & {0.7\% } & \textbf{0.3\% } & 19.5\%  & 15.5\%  \\
    4 & 15.4\%  & 6.5\%  & 15.6\%  & 15.4\%  & {0.9\% } & \textbf{0.6\% }  & 19.4\%  &  15.4\%  \\
    5 & 3.2\%  & 0.4\%  & 3.3\%  & 3.3\%  & \textbf{0.3\% } & {0.6\% }  & 3.2\%  & 3.2\%  \\
    6 & 1.5\%  & \textbf{0.1\% } & 1.6\%   & 1.5\% & 0.2\%  & {0.3\% }  & 1.5\%  & 1.5\%  \\
    7 & 1.5\%  & \textbf{0.1\% } & 1.8\%  & 1.7\%  & 0.2\%  & {0.6\% }  & 1.5\% & 1.4\%  \\
    8 & 1.5\%  & \textbf{0.1\% } & 1.7\%  & 1.4\%  & \textbf{0.1\% } & 0.2\%  & 1.5\%  & 1.6\%  \\
    9 & 1.8\%  & 0.4\%  & 1.9\%  & 1.8\%  & \textbf{0.3\% } & \textbf{0.3\% }  & 2.5\%  & 2.9\%  \\
    10 & 2.4\%  & {1.3\% } & 2.7\%  & 2.4\%  & 0.7\%  & \textbf{0.6\% }  & 1.4\%  &  1.3\%  \\
    \midrule
    Avg. gap & 5\% & 1.9\% & 5.3\% & 5.1\% & 0.4\% & 0.4\%  & 6.1\% & 5.3\% \\
    \bottomrule
    \end{tabular}}
\end{table}

Interestingly, for three instances (Instance 3, 4, and 10), DCL$^\textrm{EPL}$ finds the best policy and outperforms standard DCL. Remember that DCL$^\textrm{EPL}$ is a single policy, globally trained on all 10 problem instances. We conclude that for these instances, the \gls{EPL} embedding facilitated generalisation across different instances. Specifically, while the algorithm is trained on one set of instance parameters, it concurrently learns useful knowledge that proved effective when applied to other problem parameters. The results of \gls{EPL} are promising, as \gls{EPL} requires significantly less training time. For example, training \gls{DCL} on all seperate instances takes approximately $24$ hours on our computational platform, whereas training \gls{DCL}$^\textrm{EPL}$ for all instances simultaneously takes $5$ hours. Hence, by embedding policies in EPL, we can save significant computational effort and potentially find more robust policies.
\begin{observation}
    By training on multiple SKUs simultaneously (EPL), a learning-based approach can potentially find a robust policy that outperforms its counterpart trained separately on single SKUs.
\end{observation}
For the other instances, \gls{DCL}$^\textrm{EPL}$ finds a policy that performs the same or close to DCL. For both AVI$^\textrm{EPL}$ and PPO$^\textrm{EPL}$, we find a slightly improved policy, but the underlying weakness in both algorithms prohibits significantly better performance compared to AVI and PPO without EPL. {In Appendix~\ref{app:extra_results}, we show complementary results for which we increase the installed base from $7$ to $20$. We observe similar performance of the policies, only AVI and \gls{AVI}$^\textrm{EPL}$ significantly worsen with the larger instance sizes.}

Figure~\ref{fig:costs} depicts the costs incurred by the policies, reported per cost category for Instance 3 (\num[group-separator={,}]{10000} replications). The figure confirms that IWA, \gls{DCL}, and \gls{DCL}$^{\textrm{EPL}}$ outperform the other policies by mainly incurring fewer backorder costs.
{%
\begin{observation}
    A dual sourcing policy can save mainly in backorder costs (downtime) compared to a single sourcing policy.
\end{observation}}

\begin{figure}[hbtp]
    \centering
    \begin{tikzpicture}
\begin{axis}[
    ybar stacked,
    bar width=15pt,
    nodes near coords,
    enlargelimits=0.15,
    legend style={draw=none,at={(0.5,-0.30)},
      anchor=north,legend columns=-1},
    ylabel={Costs over episode},
    symbolic x coords={BSP, IWA, AVI, AVIEPL, \gls{DCL}, \gls{DCL}EPL, \gls{PPO}, \gls{PPO}EPL},
    xtick=data,
    xticklabels={BSP, IWA, AVI,AVI$^\textrm{EPL}$, \gls{DCL}, \gls{DCL}$^\textrm{EPL}$, \gls{PPO}, \gls{PPO}$^\textrm{EPL}$},
    x tick label style={rotate=45,anchor=east},
    point meta=explicit symbolic,
    ]
    
\addplot+[ybar,color=black,fill=black!25, opacity=0.6] plot coordinates {(BSP,47) (IWA,41.8) (AVI,48) (AVIEPL,48) (\gls{DCL},43) (\gls{DCL}EPL,42) (\gls{PPO},56) (\gls{PPO}EPL,52)};

\addplot+[ybar,color=black,fill=black!75] plot coordinates {(BSP,230) (IWA,161.8) (AVI,240.3) (AVIEPL,238) (\gls{DCL},142) (\gls{DCL}EPL,135) (\gls{PPO},270) (\gls{PPO}EPL,240)};

\addplot+[ybar,color=black,fill=black!50] plot coordinates {(BSP,120) (IWA,79.8) (AVI,122.5) (AVIEPL,120) (\gls{DCL},65.2) (\gls{DCL}EPL,62) (\gls{PPO},140) (\gls{PPO}EPL,139)};

\addplot+[ybar,color=black,fill=black,style={pattern color=black,mark=none},pattern=north west lines] plot coordinates {(BSP,390) (IWA,400.8) (AVI,391) (AVIEPL,388) (\gls{DCL},402) (\gls{DCL}EPL,399) (\gls{PPO},405) (\gls{PPO}EPL,402)};

\legend{ Holding costs, \strut Backorder costs, \strut Maintenance costs, \strut Purchase costs}

\end{axis}
\end{tikzpicture}
    \caption{Cost analysis for Instance 3.}
    \label{fig:costs}
\end{figure}
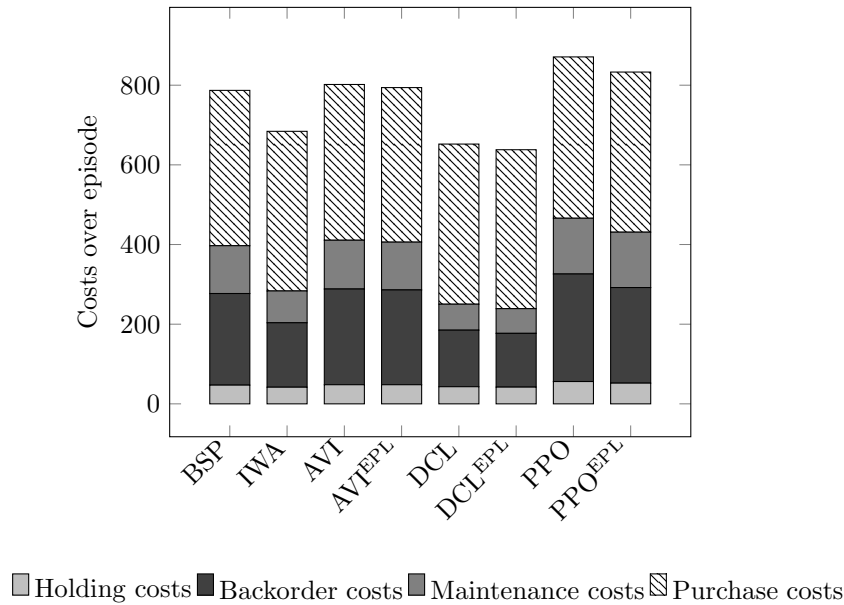

{Computation times for \gls{IWA} are below 20 seconds. \gls{AVI} requires approximately 30 seconds to train, \gls{DCL} requires between 100 and 150 minutes, and \gls{PPO} requires over 3 hours.} We note that the parameter grid $\Theta$ defined for the synthetic case is small, as there are only ten problem instances with a limited number of unique values. Hence, all SKUs have been observed during training. In a real case with many more SKUs, the parameter values are more diverse, the parameter grid is larger, and the policies are evaluated on instances that are likely not observed during training. In Section~\ref{sec:energy}, we evaluate how EPL performs in such a setting. For all remaining experiments, we focus on the best-performing policies, i.e., BSP, IWA, \gls{DCL}, and \gls{DCL}$^\textrm{EPL}$, excluding the results of the other policies.

\subsection{Energy Case}\label{sec:energy}

In this section, we explain the real-world energy sector case in Section~\ref{sect:oilgas} and show the results for this case in Section~\ref{sect:energyResult}. Our experiments are aimed at further evaluating the EPL approach and providing managerial insights.

\subsubsection{Energy Case Description}\label{sect:oilgas}

We conducted our case study using data from TotalEnergies, focusing on parts in valves used on offshore platforms in remote locations. These platforms utilise various valves, and when a valve component fails, the entire valve is replaced with a new valve and the old valve is send to a repair shop where the failed component is replaced. Then the repaired valve is send back to stock. If no replacement is available, temporary fixes can minimise production loss but still cause inconvenience and reduce efficiency, resulting in backorder costs. The components used inside these valves are available in both \gls{CM} and \gls{AM} versions.

Our dataset, summarised in Table~\ref{tab:srus}, includes five different valve types, each having several components, operating in different assets on the offshore platform. Note that because of confidentiality we do not report the absolute values but only the relative values compared to item~1. Only the values for $c_A$, $l_A$, and $m$ are ratios compared to the corresponding CM value of the same item. We performed a sensitivity analysis with various parameter settings, and added some additional instances. For instance, we also considered a larger installed base than the original data, which covers only a single platform. Hence, some parameter settings with larger installed bases have higher demand and thus can be considered fast moving. Managing spare parts for multiple platforms within a region is more efficient as we profit from the risk pooling effect, so we studied installed bases ranging from 5 to 150 valves. The single part demand is considerably lower than in the synthetic case, with mean time between failures between 3 and 20 years. {The CM lead times are longer than $8$ months.}
\begin{table}[!t]
\caption{Input parameter ratios for different spare parts compared to item~1, $c_A$, $l_A$, and $m$ are ratios of the corresponding CM parameter of the same item.}\label{tab:srus}
\centering
\resizebox{\textwidth}{!}{
\begin{tabular}{l c c c c c}
\toprule
\textbf{Input parameter} & \textbf{Item 1} & \textbf{Item 2} & \textbf{Item 3} & \textbf{Item 4} & \textbf{Item 5} \\ \midrule
No. parts per item & 3 & 2 & 6 & 1 & 4 \\
CM piece price $c_C$ ratio & 1.00 & 1.62 & 15.30 & 27.69 & 1.09 \\
CM lead time $l_C$ ratio & 1.00 & 0.98 & 0.69 & 1.09 & 0.79 \\
CM failure rate $\mu_C$ ratio & 1.00 & 1.90 & 0.31 & 1.13 & 0.28 \\
AM piece price $c_A/c_C$ ratio & 4.06 & 1.64 & 0.19 & 1.00 & 4.23 \\
AM lead time $l_A/l_C$ ratio & 0.26 & 0.31 & 0.38 & 0.36 & 0.33 \\
AM failure rate $\mu_A$ ratio & 1.00 & 1.90 & 0.31 & 1.13 & 0.28 \\
Order costs ($k_C$ and $k_A$) & 0.2$c_C$ & 0.2$c_C$ & 0.2$c_C$ & 0.2$c_C$ & 0.2$c_C$ \\
Maintenance costs $m/c_C$ ratio & 1.09 & 0.67 & 0.07 & 0.04 & 1.01 \\
Holding costs $h$ per year & 0.05$c_C$ & 0.05$c_C$ & 0.05$c_C$ & 0.05$c_C$ & 0.05$c_C$ \\
Backorder costs $b$ ratio & 1.00 & 6.11 & 63.15 & 122.64 & 8.76 \\
\bottomrule
\end{tabular}}
\end{table}
Although for this case AM parts have higher production costs than CM parts, they often have a lower failure rate, differing from our synthetic case. Backorder costs are calculated based on the inconvenience and efficiency loss during temporary fixes (a quick part repair). Backorder costs are calculated by using the currently known optimal CM stock on-hand as provided by our industry partner and determining the minimum and maximum backorder costs associated with that stock level. Next, we take the average of these two values for $b$. We will use the minimum and maximum backorder costs in the sensitivity analysis later on. Demand per item follows a Poisson process, so we do not specify $Var_C$ and $Var_A$.

Due to the high transportation costs from abroad AM facilities to offshore platforms, we now consider fixed ordering costs for both \gls{CM} and \gls{AM} parts (previously applied only to \gls{CM} parts). The parameter ranges tested in our experiments are summarised in Table~\ref{table:oilgasparameters}. Here, \enquote*{Original} indicates the parameter value provided by our industry partner. We varied five parameters: the number of offshore platforms (impacting installed base $N$), \gls{AM} purchase price, \gls{AM} lead time, \gls{AM} failure rate, and backorder costs in three different settings, resulting in $3^5=243$ unique parameter settings per part and a total of $243\cdot 5=1215$ settings across all valve types. Some settings may favor a single sourcing model, where either \gls{CM} or \gls{AM} is always optimal.

This case study is relevant as it demonstrates the practical application of \gls{AM} in the energy sector, particularly for critical components in remote locations. The high variability in demand and failure rates, coupled with significant transportation costs, highlight the potential benefits of using \gls{AM} to improve supply chain efficiency.

\begin{table}[hbtp]
\caption{Varied parameters of the energy case and their ranges.}
\centering
\begin{tabular}{l c}
\toprule
\textbf{Parameter} & \textbf{Parameter Range} \\
\midrule
No. offshore platforms & \{5, 15, 25\}\\
$c_A$ & \{Orginal, +25\%, -25\%\} \\
$l_A$ &  \{Original, -25\%, -50\%\} \\
$\mu_A$ &  \{As CM, CM-25\%, CM-50\%\} \\
b &  \{Original, Min., Max.\}\\
\bottomrule
\end{tabular}
\label{table:oilgasparameters}
\end{table}

\subsubsection{Results for the Energy Case}\label{sect:energyResult}

Note that the IWA policy is now internally using the dual index policy \citep[cf.][]{Veeraraghavan2008NowSystems}, as an exact approach (used in the synthetic case) is intractable for most energy case instances. For the EPL approach, we define a grid $|\Theta|=10^{12}$ of all problem parameters and sample from it during training, see Appendix~\ref{appendix:hyper} for details. Due to the significantly larger grid size compared to the synthetic case, EPL cannot observe all possible combinations of problem parameters during training. Note that EPL is trained once on all different items and different parameter settings to obtain a single policy.

First, we analyse the results across all $1215$ experiments, see Figure~\ref{fig:boxplots}. In $8.9\%$ of the settings, no method is able to outperform BSP, suggesting that one of the supply modes is dominant and dual sourcing is not an attractive option. In $26.6\%$ of the settings, all our proposed methods (IWA, \gls{DCL}, and \gls{DCL}$^\textrm{EPL}$) significantly outperform the \gls{BSP} baseline. In $0.7\%$ of the instances, only \gls{IWA} outperforms BSP, while in $44.7\%$ of the instances, only \gls{DCL} and \gls{DCL}$^\textrm{EPL}$ outperform BSP. For the remaining settings, no significant difference is observed between the dual sourcing policies. Figure~\ref{fig:boxplots} shows the savings made for all non-dominant settings, i.e., excluding the $8.9\%$ of settings where BSP is the best policy. The boxplot confirms that \gls{IWA} and \gls{DCL} are able to significantly improve upon the single sourcing benchmark, on average they save $16.9\%$ and $22.6\%$ in total costs compared to BSP, respectively. Furthermore, we observe that \gls{DCL} outperforms \gls{IWA} in many settings, with an average saving in total costs of $6.1\%$ compared to IWA. 
\begin{observation}
    DCL finds the overall best policy on real-world instances, saving $6.1\%$ in total costs compared to the proposed IWA heuristic and $22.6\%$ compared to the single sourcing BSP benchmark.
\end{observation}
\begin{figure}[hbtp]
    \centering
    \begin{tikzpicture}
        \begin{axis}[
            boxplot/draw direction=y,
            xtick={1, 2, 3, 4},
            xticklabels={Saving \gls{DCL} vs. BSP, Saving \gls{IWA} vs. BSP, Saving \gls{DCL} vs. IWA, Saving \gls{DCL}$^{\textrm{EPL}}$ vs. IWA},
            ylabel={Percentage (\%)},
            ymajorgrids=true,
            grid style=dashed,
            width=12cm,
            height=7cm,
            x tick label style={rotate=45,anchor=east},
            yticklabel={$\pgfmathprintnumber{\tick}\%$}
        ]
        
        \addplot+[
            boxplot prepared={
                median=27.9,
                upper quartile=28.9,
                lower quartile=18.3,
                upper whisker=31,
                lower whisker=2.9
            },
            boxplot prepared={
                average=22.6
            },
            mark=*
        ] coordinates {};

        \addplot+[
            boxplot prepared={
                median=13,
                upper quartile=28.9,
                lower quartile=9.6,
                upper whisker=37,
                lower whisker=0.0341
            },
            boxplot prepared={
                average=16.9
            },
            mark=*
        ] coordinates {};

        \addplot+[
            boxplot prepared={
                median=2.6,
                upper quartile=17.2,
                lower quartile=-0.7715,
                upper whisker=25.8366,
                lower whisker=-12.0376
            },
            boxplot prepared={
                average=6.1
            },
            mark=*
        ] coordinates {};

         \addplot+[
            boxplot prepared={
                median=1.6,
                upper quartile=9.2,
                lower quartile=-3.7715,
                upper whisker=28.8366,
                lower whisker=-13.0376
            },
            boxplot prepared={
                average=3.1
            },
            mark=*
        ] coordinates {};
        
        \end{axis}
    \end{tikzpicture}
    \caption{Boxplots of savings of the policies compared to each other.}
    \label{fig:boxplots}
\end{figure}
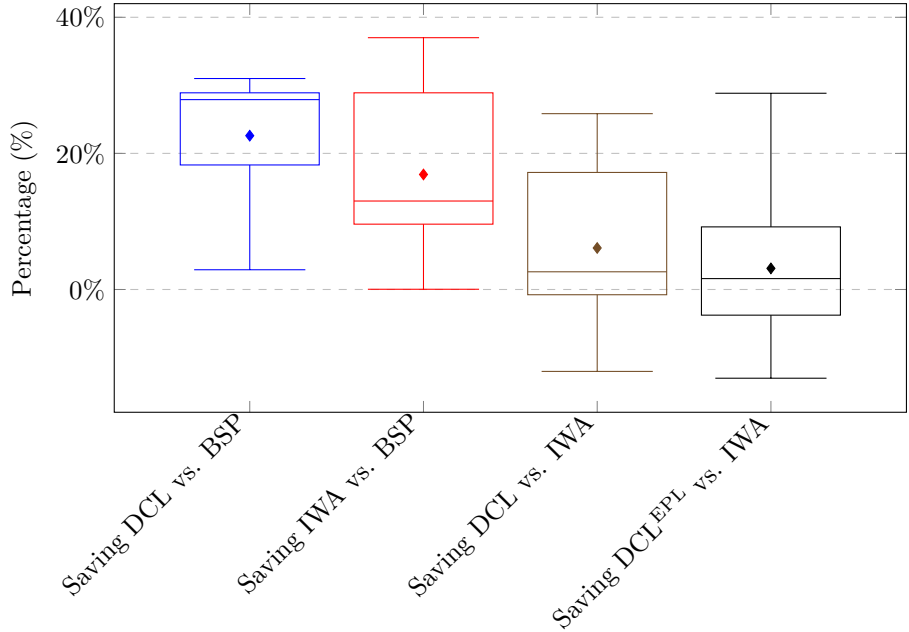

The \gls{EPL} approach (used with \gls{DCL}) is also able to outperform IWA, although less convincingly as \gls{DCL}. \gls{DCL}$^\textrm{EPL}$ saves on average $3.1\%$ in total costs compared to IWA, which is $3\%$pt worse compared to the standard DCL algorithm. However, the \gls{DCL}$^\textrm{EPL}$ is trained once on all items ($35$ hours training time), whereas the standard DCL policy is trained separately on all 1215 instances, amounting to a total training time of almost 76 days, which is a computational time saving of a factor $52$. Thus, when computational time is limited or there are thousands of different spare parts for which no separate training runs can be conducted, \gls{IWA} or \gls{DCL}$^\textrm{EPL}$ are considered preffered over DCL.
\begin{observation}
    When employing EPL -- to learn a single policy for multiple SKUs -- we achieve competitive performance while significantly reducing the computational time required for training.
\end{observation}

Next, we zoom in on the cost components and the ordering behaviour of the policies, see the barchart in Figure~\ref{fig:costsoilgas} showing costs over all non-dominated instances (\num[group-separator={,}]{10000} replications). Here, it becomes visible that \gls{DCL} is especially able to save on backorder costs compared to BSP and IWA -- by ordering larger quantities -- at the expense of higher holding costs. \gls{DCL}$^\textrm{EPL}$ seems to have learned a policy that is less able to correctly \enquote*{time} orders compared to \gls{DCL}, which results in higher backorder and purchase costs. Note that maintenance costs are low, such that these are not visible in the barchart.
{%
\begin{observation}
    For large instances, lower backorder costs (downtime) are the main reason for the savings of a dual sourcing policy compared to a single sourcing approach.
\end{observation}}

\begin{figure}[hbtp]
    \centering
    \begin{tikzpicture}
\begin{axis}[
height=6cm,
    ybar stacked,
    bar width=15pt,
    nodes near coords,
    enlargelimits=0.15,
    legend style={draw=none,at={(0.5,-0.30)},
      anchor=north,legend columns=-1},
    ylabel={Costs over episode},
    symbolic x coords={BSP, IWA, \gls{DCL}, \gls{DCL}EPL},
    xtick=data,
    xticklabels={BSP, IWA, \gls{DCL}, \gls{DCL}$^\textrm{EPL}$},
    x tick label style={rotate=45,anchor=east},
    point meta=explicit symbolic,
    ]
    
\addplot+[ybar,color=black,fill=black!25, opacity=0.6] plot coordinates {(BSP,1005) (IWA,1423) (\gls{DCL},1423) (\gls{DCL}EPL,1423)};

\addplot+[ybar,color=black,fill=black!75] plot coordinates {(BSP,34704) (IWA,22704) (\gls{DCL},1707) (\gls{DCL}EPL,7002) };

\addplot+[ybar,color=black,fill=black!50] plot coordinates {(BSP,0) (IWA,0.0) (\gls{DCL},8.3) (\gls{DCL}EPL,3.0)};

\addplot+[ybar,color=black,fill=black,style={pattern color=black,mark=none},pattern=north west lines] plot coordinates {(BSP,15515) (IWA,15515) (\gls{DCL},23991) (\gls{DCL}EPL,25000)};

\legend{ Holding costs, \strut Backorder costs, \strut Maintenance costs, \strut Purchase costs}

\end{axis}
\end{tikzpicture}
    \caption{Cost analysis for all non-dominant instances.}
    \label{fig:costsoilgas}
\end{figure}
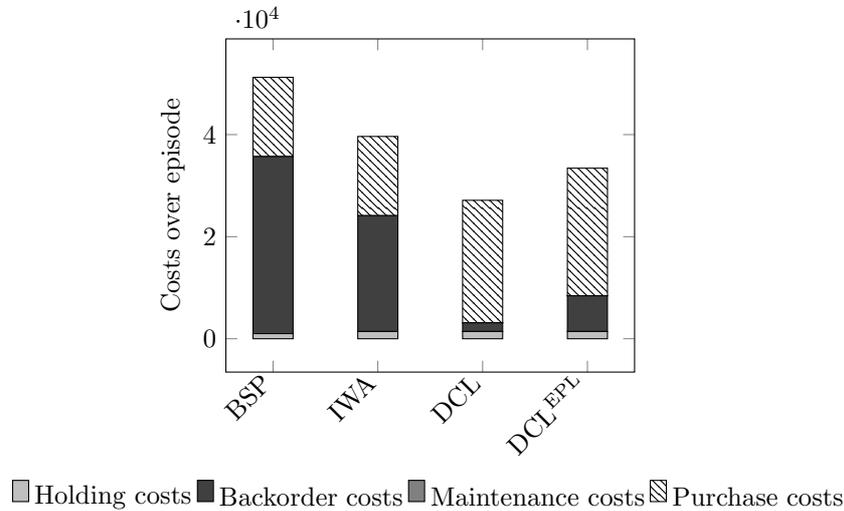

Figure~\ref{fig:scatterplot} presents a scatterplot that shows the fraction of AM parts in the installed base for the different levels of the AM mean failure rate compared to the CM failure rate: $\{$as CM, CM$-25\%$, CM$-50\%$ $\}$. The graph is constructed with the DCL policy. We separate points across the x-axis direction for visualisation purpose. The different colors indicate the sizes of the installed base: $\{\leq 25, 25-100, >100 \}$. We observe that (i) when the failure rate of AM is lower than the CM failure rate, the fraction of AM in the installed base is higher, (ii) this effect becomes smaller as the installed base size gets larger, and (iii) dual sourcing is especially relevant when the AM failure rate is low compared to CM. A few exceptions occur in case of an SKU for which AM is approximately equally expensive as CM (item 4), but yields a considerably shorter lead time. Then single sourcing using AM applies, see the dots in the upper right corner of the figure.

\begin{observation}
    When the AM failure rate is significantly lower than the CM failure rate, the fraction of AM parts in the installed base is higher, especially for smaller installed bases. This effect diminishes with larger installed bases, with some exceptions when the AM and CM piece prices are similar.
\end{observation}

\begin{figure}[!b]
    \centering
    \begin{tikzpicture}
        \begin{axis}[
            width=0.95\textwidth,
            height = 6.5cm,
                 xlabel={Mean failure rate AM},
    ylabel={Fraction AM in installed base},
    legend style={draw=none,at={(0.5,-0.33)}, anchor=north, legend columns=-1},
    xtick = {-0.5,-0.25,0},
    xticklabels = {CM-50\%, CM-25\%,as CM},
    scatter/classes={
            red={mark=*,draw=red,fill=red, mark size=2pt, opacity=1.0},    % Smaller, semi-transparent circles for 'low'
            blue={mark=square*,draw=green,fill=green, mark size=2pt, opacity=1.0}, % Smaller, semi-transparent squares for 'medium'
            green={mark=triangle*,draw=blue,fill=blue, mark size=2pt}  % Smaller, semi-transparent triangles for 'high'
        }
]

\addplot[
    scatter,
    only marks,
    scatter src=explicit symbolic
] table [meta=label] {
    x y label
0.002	0.992162373	red
0.009    0.992162373	red
0.0005  0.992162373	red
0.007   0.992162373	red
0.0000234   0.992162373	red
0.00002 0.992162373	blue
-0.495000837	0.21714262	red
-0.493360315	0.216859665	red
-0.493965489	0.216268693	red
-0.2475117	0.063589473	red
-0.249237929	0.063270258	red
-0.253424219	0.063306151	red
-0.001647387	0.03194892	red
0.00511149	0.031452232	red
4.06E-05	0.031185344	red
-0.495979518	0.216310421	red
-0.506846902	0.218608904	red
-0.497087322	0.218091626	red
-0.253656516	0.062965716	red
-0.235876379	0.031396273	red
-0.24678289	0.062777741	red
0.004048375	0.031473562	red
0.001138576	0.03157833	red
-0.001512153	0.032211403	red
-0.49995138	0.212578787	red
-0.502316431	0.217471682	red
-0.504901101	0.216823073	red
-0.247827239	0.063797298	red
-0.255138495	0.06164141	red
-0.249262454	0.061647405	red
-0.001024381	0.031956869	red
0.001690102	0.031683413	red
-0.004535299	0.03203114	red
-0.508688191	0.214337319	red
-0.500759667	0.212215681	red
-0.506252622	0.212312502	red
-0.24630519	0.062270009	red
-0.249136998	0.062963558	red
-0.251839956	0.063372606	red
-0.00759016	0.031757146	red
0.004866243	0.031894975	red
0.00906701	0.032353528	red
-0.487618215	0.063562261	red
-0.501565205	0.217240261	red
-0.498630839	0.214992882	red
-0.242814959	0.063573919	red
-0.247272742	0.06186696	red
-0.252595676	0.063202192	red
-0.003901848	0.031445608	red
0.002592445	0.03163452	red
-0.003858669	0.031353395	red
-0.496370957	0.215421791	red
-0.502235749	0.213599267	red
-0.497922873	0.216286558	red
-0.252901665	0.063340189	red
-0.2490673	0.063835137	red
-0.245567224	0.064339304	red
0.000574641	0.031476384	red
-0.000183499	0.031052452	red
0.004354391	0.031564226	red
-0.505532888	0.213864091	red
-0.496264558	0.210132775	red
-0.505425976	0.21351153	red
-0.251183278	0.063448486	red
-0.246903969	0.062067097	red
-0.241911842	0.0637762	red
-0.000625017	0.032035131	red
-0.001860964	0.031810161	red
0.003345398	0.031695773	red
-0.503320375	0.209852458	red
-0.491701954	0.212886592	red
-0.504136535	0.219022641	red
-0.251055386	0.062391653	red
-0.246757066	0.063544794	red
-0.248204323	0.062933599	red
0.005118845	0.031663999	red
-0.00593759	0.031647407	red
-0.002720473	0.031468714	red
-0.50218002	0.210914478	red
-0.50317636	0.2150992	red
-0.505885323	0.213654062	red
-0.252216287	0.063804836	red
-0.25849165	0.063100656	red
-0.246201452	0.063267272	red
0.000476498	0.031923511	red
-0.005966787	0.031422792	red
-7.58E-05	0.031527078	red
-0.503330071	0.217602985	red
-0.506703846	0.213716286	red
-0.492254808	0.213088211	red
-0.248933192	0.063845896	red
-0.25278106	0.063983896	red
-0.25908157	0.064166881	red
-0.003319293	0.031889399	red
0.0084777	0.031639745	red
0.001344509	0.031946365	red
-0.50048017	0.215138627	red
-0.497304022	0.214165985	red
-0.503572438	0.209090961	red
-0.254737486	0.061791732	red
-0.259365008	0.063250814	red
-0.263580375	0.063638899	red
-0.007479304	0.032360533	red
-0.004631329	0.031096941	red
-0.006521453	0.031637372	red
-0.502874996	0.217914835	red
-0.499770812	0.214461799	red
-0.503497287	0.213505067	red
-0.248940879	0.063382991	red
-0.259517121	0.06369808	red
-0.240763057	0.031413716	red
-0.000644047	0.031415515	red
-0.000897526	0.031950029	red
0.006350434	0.031676334	red
-0.49961617	0.216819905	red
-0.495549738	0.215105435	red
-0.500809201	0.213837288	red
-0.251673248	0.06352934	red
-0.245532847	0.062798613	red
-0.248009955	0.06377258	red
-0.006262164	0.030823047	red
0.010772759	0.03044945	red
0.003308064	0.031053716	red
-0.496642483	0.212377585	red
-0.496707165	0.214001384	red
-0.503830025	0.21258809	red
-0.257331117	0.063790171	red
-0.253782109	0.062974306	red
-0.249657695	0.062687519	red
-0.004603854	0.031267947	red
-0.005904705	0.031329995	red
0.000229299	0.032237905	red
-0.496470885	0.219610351	red
-0.49629446	0.213259377	red
-0.504042085	0.216117143	red
-0.242963434	0.061926014	red
-0.248794775	0.063209373	red
-0.259780779	0.062341717	red
0.007684649	0.03116237	red
0.005044621	0.031759993	red
0.001091648	0.031434487	red
-0.506616384	0.2146261	red
-0.495612916	0.214202792	red
-0.507573349	0.215796791	red
-0.246501807	0.064411017	red
-0.243356236	0.06293425	red
-0.255949136	0.062947345	red
0.001955197	0.031532536	red
0.005027091	0.031754796	red
0.000293205	0.031353772	red
-0.496447829	0.213851315	red
-0.495165558	0.21648483	red
-0.494133422	0.213064439	red
-0.250300092	0.063699138	red
-0.244443541	0.063577027	red
-0.242455052	0.062350855	red
-0.006161323	0.031829257	red
-0.000941696	0.03132467	red
-0.001037625	0.030830942	red
-0.499495598	0.215454111	red
-0.507743592	0.218651939	red
-0.50115957	0.212605183	red
-0.245551764	0.063381718	red
-0.246168818	0.063205023	red
-0.24714427	0.063472258	red
-0.003660861	0.031738879	red
-0.007255439	0.031572196	red
-0.004748209	0.031445064	red
-0.498130885	0.214732441	red
-0.501012669	0.211221355	red
-0.495419023	0.215202645	red
-0.245263915	0.063013838	red
-0.258150764	0.061996977	red
-0.25013733	0.062003177	red
0.003379944	0.031194417	red
0.000551922	0.031950761	red
-0.002775883	0.030836437	red
-0.506330949	0.220982053	red
-0.509196731	0.214778137	red
-0.49700323	0.212920466	red
-0.256199061	0.064251619	red
-0.24786648	0.063134097	red
-0.256445705	0.064502	red
-0.005339841	0.031453117	red
-0.006127752	0.031182913	red
0.003700667	0.03160623	red
-0.512410583	0.210566579	red
-0.498269604	0.214761612	red
-0.509929305	0.214212076	red
-0.245605181	0.063617175	red
-0.254531541	0.062616675	red
-0.243543537	0.06238374	red
-0.011945039	0.03198589	red
0.003158842	0.032006044	red
-0.003574538	0.031920452	red
-0.506930276	0.214015963	red
-0.507990954	0.21378157	red
-0.496365488	0.215255828	red
-0.251194406	0.063169135	red
-0.24838766	0.063527794	red
-0.247650991	0.063676284	red
-0.003487093	0.031409079	red
0.004585902	0.03164347	red
0.003665887	0.031305712	red
-0.504745336	0.210711672	red
-0.513946174	0.218652161	red
-0.496319517	0.215359826	red
-0.25654251	0.064048662	red
-0.254834429	0.064164979	red
-0.255314755	0.063224364	red
0.000126506	0.030802455	red
-0.001123701	0.03187658	red
-0.006105148	0.031677659	red
-0.503695716	0.214980482	red
-0.504004979	0.214573863	red
-0.494416673	0.21559815	red
-0.255603093	0.063900361	red
-0.247439268	0.063085888	red
-0.257317184	0.06441264	red
0.001993624	0.031417777	red
-0.003673754	0.031235223	red
-0.002819492	0.031438881	red
-0.505783766	0.220044622	red
-0.514059172	0.21598331	red
-0.504150094	0.212423148	red
-0.250404726	0.062071264	red
-0.247243609	0.061941551	red
-0.247257849	0.063477759	red
-0.004829988	0.03154398	red
-0.004555792	0.031525358	red
-0.005058934	0.031279664	red
-0.508296989	0.214220743	red
-0.505030683	0.214935883	red
-0.496333912	0.217157963	red
-0.252304876	0.062808483	red
-0.248158093	0.063185669	red
-0.251964448	0.062358243	red
0.004339447	0.031541304	red
0.006307883	0.032253604	red
-0.008251275	0.03106684	red
-0.502260213	0.213723462	red
-0.50967729	0.215497151	red
-0.506436999	0.216697261	red
-0.244963048	0.063020093	red
-0.248876114	0.062391484	red
-0.245985219	0.062573363	red
-0.000866501	0.031635839	red
-0.00557887	0.031990852	red
-0.001609352	0.031271124	red
-0.495609338	0.215600095	red
-0.50522073	0.218547596	red
-0.494829778	0.214106524	red
-0.251267143	0.064064947	red
-0.246977282	0.062339815	red
-0.258965394	0.062445217	red
-0.000205827	0.030857484	red
0.000370922	0.031696076	red
0.00114108	0.031053775	red
-0.490350149	0.210462036	red
-0.498884881	0.213343513	red
-0.498826276	0.213443079	red
-0.241986149	0.063212618	red
-0.246407161	0.063598702	red
-0.237532904	0.031812487	red
0.001799972	0.032131913	red
-0.000494458	0.03117456	red
0.001309263	0.031159627	red
-0.502921536	0.208564127	red
-0.512247599	0.211287192	red
-0.496210451	0.212622119	red
-0.25455035	0.063758291	red
-0.243032822	0.061503118	red
-0.255470155	0.062249575	red
0.002401448	0.031625901	red
-0.007097233	0.031337876	red
-0.000722843	0.031095963	red
-0.503802509	0.21774967	red
-0.500528778	0.214459535	red
-0.489616921	0.062892199	red
-0.2449556	0.064639391	red
-0.246416462	0.062663034	red
-0.254218164	0.063660311	red
-0.004758412	0.031799541	red
0.01533303	0.031284593	red
0.009579625	0.031155216	red
-0.499233565	0.211479058	red
-0.500916986	0.213595571	red
-0.505020454	0.216681536	red
-0.24497195	0.062611272	red
-0.249168214	0.062998538	red
-0.252697063	0.063706358	red
-0.005040778	0.031259612	red
-0.005742334	0.032270154	red
0.006429588	0.030666839	red
-0.496781122	0.215982636	red
-0.503583105	0.215459884	red
-0.495269282	0.215748463	red
-0.247026945	0.062753096	red
-0.247571677	0.06209466	red
-0.245619629	0.063292705	red
-0.002275092	0.03143426	red
-0.001793944	0.030985997	red
-0.001721986	0.031321089	red
-0.503414966	0.212978419	red
-0.50316986	0.214400965	red
-0.504632335	0.210999276	red
-0.24573439	0.063735972	red
-0.254189607	0.063367054	red
-0.252222641	0.062759473	red
-0.002588583	0.031658915	red
-0.002232684	0.031653244	red
-0.010487537	0.03204067	red
-0.500348547	0.214329116	red
-0.500833011	0.210978525	red
-0.501128475	0.214847335	red
-0.242036008	0.063033355	red
-0.248818451	0.064292699	red
-0.258552663	0.064769567	red
0.009646391	0.031420775	red
0.011249074	0.031692686	red
0.004360422	0.031266195	red
-0.49600687	0.217317385	red
-0.501350214	0.218070976	red
-0.492690114	0.212429397	red
-0.256520329	0.063351295	red
-0.25336615	0.063412041	red
-0.233083805	0.031537661	red
0.004939751	0.031653816	red
0.006148876	0.031550777	red
0.002767506	0.030923223	red
-0.495068053	0.208022941	red
-0.491935801	0.206223696	red
-0.495805557	0.210651299	red
-0.250323317	0.061916159	red
-0.242058793	0.061474063	red
-0.25254436	0.061920758	red
-0.000307396	0.030023039	red
-0.002289453	0.030907128	red
-0.00013226	0.030858719	red
-0.497963034	0.209479119	red
-0.501764969	0.209636945	red
-0.501524303	0.210676367	red
-0.250503693	0.059644288	red
-0.245966175	0.061853738	red
-0.248223915	0.062385324	red
0.002635386	0.030890789	red
-0.005899675	0.030761584	red
-0.012234674	0.030971982	red
-0.508641995	0.211507738	red
-0.499702998	0.210772936	red
-0.499083207	0.210711134	red
-0.251481159	0.061443495	red
-0.256638957	0.062260411	red
-0.25442268	0.061364456	red
-0.00202397	0.030144358	red
-0.000387207	0.030890941	red
0.003481324	0.030852384	red
-0.49907301	0.20800017	red
-0.506722007	0.206050617	red
-0.501641969	0.212733002	red
-0.253426605	0.061287299	red
-0.262766071	0.061772234	red
-0.255881812	0.061492583	red
0.00568096	0.030778757	red
0.005587774	0.030972713	red
-0.001003137	0.030735821	red
-0.5016679	0.211545324	red
-0.492983211	0.212416726	red
-0.503826366	0.211506425	red
-0.249916179	0.061331963	red
-0.250623301	0.062779686	red
-0.247020845	0.062787235	red
0.003794892	0.030047765	red
-0.000919068	0.030980531	red
0.001725753	0.031420065	red
-0.500250932	0.209338716	red
-0.498334607	0.207950158	red
-0.495804927	0.212171515	red
-0.246495328	0.061260304	red
-0.254830362	0.062711706	red
-0.25243641	0.062304233	red
0.0033307	0.030653243	red
0.003496785	0.031636751	red
0.005208453	0.031268619	red
-0.509295492	0.211804676	red
-0.505733308	0.211392512	red
-0.502465115	0.213647006	red
-0.242930564	0.060894041	red
-0.246094361	0.061805298	red
-0.244059781	0.063290155	red
-0.004806414	0.03055337	red
-0.004287749	0.03074263	red
-0.001853734	0.030508323	red
-0.502365012	0.207067695	red
-0.494717008	0.21375074	red
-0.495949522	0.207978334	red
-0.240493073	0.030654267	red
-0.246378426	0.061342124	red
-0.240884464	0.061729888	red
-0.006217583	0.030791515	red
-0.003948834	0.030444207	red
0.005029407	0.030882922	red
-0.502836449	0.20988347	red
-0.490327432	0.204045576	red
-0.500852176	0.211590742	red
-0.241629873	0.061449172	red
-0.261676577	0.060510314	red
-0.245210171	0.061813802	red
-0.00212109	0.030622579	red
0.003834324	0.031448949	red
-0.001658795	0.031338057	red
-0.501120487	0.200776516	red
-0.497014485	0.198804238	red
-0.499504635	0.203139605	red
-0.24620732	0.058051481	red
-0.258431816	0.057369641	red
-0.254211186	0.058900459	red
-0.000647189	0.029069843	red
-0.00884014	0.029508943	red
-0.005376154	0.029520884	red
-0.499303564	0.202699502	red
-0.501120945	0.199763303	red
-0.495876092	0.199359331	red
-0.248169737	0.059924925	red
-0.255021077	0.059724518	red
-0.249506916	0.059770336	red
0.002024421	0.030314909	red
-0.001382977	0.029136631	red
0.004717627	0.029655753	red
-0.49657368	0.201551337	red
-0.494598721	0.199589711	red
-0.502139087	0.201194939	red
-0.242752886	0.058123066	red
-0.248939045	0.058607349	red
-0.255932457	0.058774097	red
0.001365513	0.029775204	red
0.005157534	0.029407475	red
0.002429968	0.029909202	red
-0.512151183	0.19852263	red
-0.493658241	0.198542627	red
-0.507022857	0.200934922	red
-0.257758097	0.058627378	red
-0.255946602	0.058919038	red
-0.256261065	0.059457951	red
-0.00461454	0.028734748	red
0.00042024	0.030345675	red
0.00969014	0.028539345	red
-0.495174153	0.197749966	red
-0.504239239	0.200895296	red
-0.500739516	0.201583797	red
-0.247196692	0.059333781	red
-0.245264659	0.0592425	red
-0.242019369	0.058984375	red
0.011587113	0.02976264	red
-0.009711638	0.029528344	red
0.001404611	0.030356572	red
-0.502979925	0.203792248	red
-0.50336371	0.200721186	red
-0.497664655	0.197009384	red
-0.257468336	0.058650898	red
-0.252027195	0.058739425	red
-0.251178112	0.058793744	red
-0.000208391	0.029799926	red
0.003877338	0.029891649	red
-0.002363415	0.029632324	red
-0.499249959	0.201152864	red
-0.495353324	0.200527789	red
-0.504016129	0.202607981	red
-0.24730479	0.058797389	red
-0.24935927	0.059434494	red
-0.245277713	0.058395221	red
-0.003373549	0.030009348	red
0.008270975	0.028937558	red
-0.002471236	0.029340588	red
-0.503854817	0.20088426	red
-0.493880487	0.203961903	red
-0.501979429	0.203452079	red
-0.241910597	0.059481016	red
-0.248320307	0.058860199	red
-0.25822041	0.058994038	red
0.002574887	0.029985047	red
-0.003546243	0.029892228	red
-0.00588369	0.029406279	red
-0.496524097	0.206692752	red
-0.507174503	0.202108933	red
-0.496717782	0.198255076	red
-0.245771525	0.059807451	red
-0.251193212	0.059786251	red
-0.252823067	0.058664006	red
-0.010599843	0.029781875	red
-0.008727985	0.029811804	red
-0.004813905	0.02958949	red
-0.503976319	0.992162373	red
-0.503075055	0.155523875	red
-0.496893232	0.958928693	red
-0.254534387	0.044376465	red
-0.249945829	0.044623141	red
-0.252479553	0.044532684	red
0.004852054	0.022890341	red
0.001688305	0.022621411	red
-0.000495456	0.022613753	red
-0.512188557	0.149567638	red
-0.503192124	0.154926232	red
-0.509014414	0.150807904	red
-0.242361414	0.044667916	red
-0.25888754	0.284413957	red
-0.251725749	0.287219853	red
-0.003745428	0.023018445	red
-0.002346553	0.022499749	red
-0.007031233	0.022610765	red
-0.499703681	0.157648091	red
-0.499409771	0.149838571	red
-0.497281076	0.975164069	red
-0.244246526	0.043827141	red
-0.255402419	0.045195846	red
-0.252297604	0.283372765	red
8.81E-05	0.022078693	red
0.006259214	0.022578041	red
0.010863333	0.143527476	red
-0.490744998	0.15025111	red
-0.493993	0.153921319	red
-0.502699551	0.148857642	red
-0.246714561	0.044114815	red
-0.250983648	0.045942618	red
-0.243412922	0.044383334	red
-0.000872541	0.022793109	red
0.00807751	0.022946782	red
0.002755594	0.022459653	red
-0.514877888	0.151467217	red
-0.498925334	0.151410935	red
-0.495395285	0.152122843	red
-0.256570429	0.043190714	red
-0.252289199	0.044615653	red
-0.246447613	0.044267492	red
0.000716633	0.022325867	red
0.002118386	0.022498546	red
0.010528809	0.142864317	red
-0.495525428	0.147570964	red
-0.498654608	0.153580611	red
-0.498739315	0.151893742	red
-0.250087081	0.045510445	red
-0.245677905	0.044306569	red
-0.252856924	0.287147694	red
-0.007591809	0.022096498	red
0.003682608	0.022921144	red
-0.003403784	0.022239764	red
-0.496309742	0.151982908	red
-0.506775561	0.152058807	red
-0.500843064	0.156359721	red
-0.245513782	0.044206996	red
-0.253492643	0.044126125	red
-0.251191974	0.260615776	red
0.008668607	0.022206194	red
-0.005466361	0.022025833	red
-0.004300281	0.022338929	red
-0.501022012	0.150407697	red
-0.497664901	0.15236849	red
-0.497591844	0.152913095	red
-0.251926232	0.044029659	red
-0.254633706	0.043915326	red
-0.255422284	0.044489902	red
0.016579814	0.143940146	red
0.004127954	0.022499671	red
-0.010813809	0.022314053	red
-0.500851251	0.152045975	red
-0.497198219	0.151099988	red
-0.493214826	0.96620329	red
-0.250927943	0.046141291	red
-0.251100973	0.046001046	red
-0.247192133	0.044854487	red
0.000317771	0.022054412	red
-0.002323754	0.144704063	red
0.000259635	0.02259482	red
-0.501839238	0.198742046	blue
-0.507832152	0.19086249	blue
-0.499738774	0.197350888	blue
-0.255051542	0.05737461	blue
-0.254154391	0.057328487	blue
-0.253876144	0.057635796	blue
-0.004770667	0.028713035	blue
-0.003595516	0.028563309	blue
-0.001029931	0.029887466	blue
-0.501995166	0.200155103	blue
-0.507729335	0.19399443	blue
-0.503803823	0.195758813	blue
-0.254107882	0.057623514	blue
-0.246668091	0.057867618	blue
-0.262688672	0.057540913	blue
-0.003747145	0.028865254	blue
-0.005518413	0.028667171	blue
-0.004464714	0.028778231	blue
-0.494310016	0.195823608	blue
-0.507216248	0.195671506	blue
-0.501894404	0.191310005	blue
-0.252188107	0.05596443	blue
-0.250374955	0.057967792	blue
-0.260164947	0.057905911	blue
0.000766275	0.028681029	blue
-4.08E-05	0.02890352	blue
0.000803008	0.028553455	blue
-0.505055345	0.200595589	blue
-0.501303484	0.19665583	blue
-0.502599302	0.199788674	blue
-0.250505374	0.05825704	blue
-0.244798947	0.057717883	blue
-0.256045839	0.059130587	blue
-0.011910476	0.029094721	blue
-0.006043128	0.028919199	blue
-0.002725168	0.029010922	blue
-0.506328444	0.200302061	blue
-0.494744944	0.198876598	blue
-0.49427533	0.19243757	blue
-0.258207739	0.05757088	blue
-0.255846688	0.057243038	blue
-0.246989987	0.057934952	blue
-0.001682312	0.028708626	blue
0.005479841	0.028574334	blue
-0.003243157	0.028579236	blue
-0.49805239	0.193109825	blue
-0.504344022	0.195661366	blue
-0.512064041	0.196634469	blue
-0.247461679	0.057696095	blue
-0.25094086	0.058061854	blue
-0.247825147	0.056573709	blue
-0.008334995	0.029072044	blue
-0.011060065	0.028470639	blue
-0.001737828	0.028275469	blue
-0.490612596	0.198439339	blue
-0.497175235	0.195624416	blue
-0.510966472	0.196400854	blue
-0.246010873	0.058751143	blue
-0.240853556	0.028751859	blue
-0.250928994	0.056860013	blue
0.002386191	0.028941348	blue
-0.011144432	0.028810914	blue
-0.000812409	0.029393851	blue
-0.501951362	0.194226744	blue
-0.50583099	0.197513011	blue
-0.505146956	0.199010259	blue
-0.249810365	0.058578903	blue
-0.248660403	0.05728809	blue
-0.251941497	0.057640975	blue
-0.004397041	0.028748603	blue
0.006804773	0.029759027	blue
0.00543621	0.028797765	blue
-0.497585169	0.198259555	blue
-0.503234299	0.194465833	blue
-0.51193902	0.194943816	blue
-0.254295966	0.058245664	blue
-0.25371344	0.058198696	blue
-0.258493857	0.057189445	blue
0.003138464	0.029083745	blue
-0.00232228	0.028870628	blue
-0.005292387	0.028533573	blue
-0.503890525	0.211350195	red
-0.499241059	0.21875782	red
-0.502664922	0.21600218	red
-0.251102725	0.064055172	red
-0.253572832	0.063177539	red
-0.244828258	0.062732372	red
0.001789111	0.031631618	red
-0.004553045	0.031922756	red
-0.000814666	0.031280738	red
-0.497383663	0.21528753	red
-0.498104993	0.215971615	red
-0.486478705	0.062820617	red
-0.248807431	0.063672376	red
-0.247413492	0.061420543	red
-0.2500947	0.062725258	red
0.006068783	0.031805611	red
-0.003189288	0.03112823	red
0.000839324	0.031055538	red
-0.49910069	0.212513051	red
-0.493152224	0.217689964	red
-0.504461017	0.208187173	red
-0.24936297	0.064266921	red
-0.251334304	0.063783495	red
-0.248123119	0.06273769	red
0.008834073	0.032408526	red
-0.003558813	0.031896294	red
0.004454115	0.031560966	red
-0.497522387	0.219923807	red
-0.502137526	0.2125139	red
-0.508241145	0.212438603	red
-0.242473609	0.063097561	red
-0.248453324	0.06325821	red
-0.25575651	0.061847595	red
-0.003115481	0.031804616	red
-0.003792499	0.031378876	red
-0.00596909	0.031634565	red
-0.504634119	0.214454838	red
-0.508586812	0.218145709	red
-0.503786808	0.208986296	red
-0.249427379	0.062601003	red
-0.242892757	0.062836069	red
-0.250973354	0.062675879	red
0.001096528	0.030981352	red
0.004653575	0.031375615	red
0.001127853	0.031893891	red
-0.492450069	0.21309393	red
-0.493258167	0.216461598	red
-0.497529616	0.215199214	red
-0.249302202	0.062922116	red
-0.249194789	0.063136985	red
-0.256213151	0.064110726	red
-0.006303652	0.031848036	red
0.001080503	0.03160372	red
0.004607133	0.032143637	red
-0.496415821	0.214166826	red
-0.497455796	0.213152914	red
-0.496346617	0.212700649	red
-0.252578016	0.062213236	red
-0.250067432	0.062416887	red
-0.250502328	0.06447572	red
-0.00031307	0.031383442	red
-0.004227569	0.031166759	red
0.001183706	0.031526369	red
-0.503540252	0.213033229	red
-0.505178104	0.214535379	red
-0.502368709	0.212076771	red
-0.24627933	0.063962028	red
-0.242381953	0.063128795	red
-0.252139943	0.062761636	red
0.001357961	0.031418476	red
0.001391355	0.031640116	red
0.001884866	0.03151035	red
-0.49211034	0.217582732	red
-0.497985015	0.214216871	red
-0.500615862	0.214308297	red
-0.245484908	0.062742546	red
-0.248821986	0.063137074	red
-0.255840281	0.063579517	red
0.00749175	0.031280804	red
-0.009296726	0.031279026	red
0.005975098	0.031360616	red
-0.496377662	0.105380254	blue
-0.504558077	0.492202904	blue
-0.502201325	0.436308376	blue
-0.245809222	0.028808267	blue
-0.24741349	0.145532263	blue
-0.248143614	0.129088243	blue
-0.004239508	0.014674558	blue
-0.00377712	0.072920463	blue
0.003105751	0.064797148	blue
-0.497211025	0.104349499	blue
-0.504381932	0.488693979	blue
-0.505427502	0.441027775	blue
-0.249325513	0.030090806	blue
-0.25660311	0.145030446	blue
-0.242996063	0.128186297	blue
0.004262682	0.014956651	blue
-0.000759701	0.071969426	blue
-0.003735824	0.06428849	blue
-0.502689852	0.099729443	blue
-0.500337629	0.484916786	blue
-0.502950395	0.428558516	blue
-0.246627429	0.030841389	blue
-0.247667473	0.144217493	blue
-0.246420108	0.126310728	blue
-0.002688621	0.015138677	blue
-0.005047264	0.071742935	blue
0.006634731	0.062509949	blue
-0.506941193	0.096990183	blue
-0.495217988	0.497498167	blue
-0.497684666	0.437840032	blue
-0.259952456	0.030579819	blue
-0.243705287	0.14374347	blue
-0.252823738	0.130237156	blue
0.000402096	0.014070331	blue
0.001471045	0.072350329	blue
0.000838534	0.064304705	blue
-0.497976287	0.099748299	blue
-0.497447562	0.494818032	blue
-0.502203252	0.433220304	blue
-0.251396075	0.030659391	blue
-0.248782458	0.144774312	blue
-0.250853676	0.129224159	blue
-0.003003735	0.014682216	blue
0.005782626	0.072506643	blue
-0.000648329	0.064248835	blue
-0.499909944	0.106291414	blue
-0.501790561	0.494457877	blue
-0.507131715	0.427985876	blue
-0.249063853	0.030570799	blue
-0.253640291	0.144233906	blue
-0.247857741	0.127425814	blue
-0.00444784	0.014606852	blue
0.0025403	0.072403925	blue
0.001537583	0.063259398	blue
-0.501894453	0.10059143	blue
-0.503962296	0.494237127	blue
-0.501137283	0.439743589	blue
-0.249378818	0.029878315	blue
-0.250137229	0.144639745	blue
-0.254642852	0.12903384	blue
0.004170055	0.015108681	blue
-0.000396353	0.072359803	blue
0.001198947	0.064182612	blue
-0.493398764	0.103677157	blue
-0.49881282	0.493090766	blue
-0.495524399	0.433427296	blue
-0.250082793	0.029829629	blue
-0.241108454	0.144079863	blue
-0.255663042	0.128078801	blue
0.007502539	0.015397785	blue
-0.008404622	0.072493286	blue
-0.00483331	0.064265472	blue
-0.503643065	0.106705733	blue
-0.492094679	0.489360194	blue
-0.501510979	0.432539849	blue
-0.234870639	0.014393138	blue
-0.249933953	0.143679465	blue
-0.241581148	0.12659687	blue
0.001558918	0.015509349	blue
-0.010329749	0.072685804	blue
0.001819641	0.063007089	blue
-0.492005464	0.1154749	blue
-0.496361576	0.60330734	blue
-0.503836858	0.117749575	blue
-0.250456258	0.033683628	blue
-0.248606733	0.034298693	blue
-0.253033344	0.034476685	blue
-0.008125786	0.016951029	blue
0.004265211	0.017087536	blue
-0.009856163	0.017735159	blue
-0.504944689	0.614057874	blue
-0.495851581	0.522351182	blue
-0.504013292	0.11638237	blue
-0.245072611	0.158351227	blue
-0.247612458	0.034009614	blue
-0.244048591	0.034321686	blue
0.005430354	0.089677323	blue
0.000325422	0.08910802	blue
-0.004791226	0.017631966	blue
-0.498522759	0.115075751	blue
-0.510655855	0.603172188	blue
-0.499807465	0.603689852	blue
-0.254158652	0.031864843	blue
-0.24960919	0.034224402	blue
-0.248490857	0.034395653	blue
0.007184748	0.090039126	blue
0.00229945	0.01696164	blue
0.004047459	0.017052369	blue
-0.498225825	0.120067755	blue
-0.499518178	0.118572658	blue
-0.503176757	0.116769664	blue
-0.255095177	0.034046929	blue
-0.25253731	0.177718018	blue
-0.241334469	0.035130733	blue
-0.001724414	0.017066113	blue
-0.004741768	0.017813997	blue
-0.003799787	0.089287035	blue
-0.495304737	0.610543445	blue
-0.491856967	0.117912073	blue
-0.50503665	0.116102444	blue
-0.249361454	0.174161487	blue
-0.246842378	0.033890815	blue
-0.253357119	0.032150397	blue
-0.006028411	0.089678557	blue
-0.004052321	0.016992409	blue
-0.006652602	0.017665167	blue
-0.497007616	0.118216277	blue
-0.49491644	0.606688136	blue
-0.50253271	0.11693451	blue
-0.256008654	0.033593891	blue
-0.258816593	0.035329316	blue
-0.247847387	0.035042397	blue
0.002740857	0.017245881	blue
-0.000801815	0.017068911	blue
-0.008405431	0.016933781	blue
-0.501979128	0.119518132	blue
-0.496693631	0.12035727	blue
-0.501186563	0.583405507	blue
-0.249164648	0.034140209	blue
-0.240601978	0.017147774	blue
-0.255986327	0.171520066	blue
-0.001831836	0.016947157	blue
0.004764297	0.016796274	blue
0.003424112	0.018133562	blue
-0.502207721	0.116763687	blue
-0.504047849	0.549399314	blue
-0.505939917	0.121180802	blue
-0.249932343	0.174615733	blue
-0.252330918	0.034840999	blue
-0.254397768	0.177858563	blue
0.005383464	0.017222681	blue
0.004590187	0.086785488	blue
-0.00176312	0.01734201	blue
-0.504754765	0.121914376	blue
-0.494206385	0.583744048	blue
-0.501323884	0.117114992	blue
-0.247448572	0.035620563	blue
-0.250684042	0.03480261	blue
-0.248675289	0.035115443	blue
-0.003752757	0.089363473	blue
0.006801076	0.017721318	blue
0.002721098	0.086391523	blue
-0.495058593	0.006396276	red
-0.499367701	0.011622506	red
-0.498227836	0.024791813	red
-0.257109131	0.183590051	red
-0.247220495	0.005567475	red
-0.251659494	0.006960107	red
0.009098622	0.000656255	red
0.006457107	0.085988706	red
0.005284751	0.085310688	red
-0.496829836	0.001813983	red
-0.505970068	0.012409307	red
-0.511164703	0.024012637	red
-0.248348159	0.001868788	red
-0.24912193	0.175147128	red
-0.243377322	0.171999201	red
-0.002885831	0.000641519	red
-0.004561765	0.001690862	red
-0.00092671	0.078312055	red
-0.502618026	0.006268044	red
-0.50031718	0.606326344	red
-0.495288479	0.027647088	red
-0.250288326	0.002098945	red
-0.250040239	0.004340182	red
-0.250080491	0.167278849	red
0.007727633	0.090946054	red
-0.003332436	0.088163843	red
-0.000576874	0.080385495	red
-0.504647067	0.008111059	red
-0.506739172	0.008865198	red
-0.504863962	0.03008648	red
-0.249191369	0.000862971	red
-0.244071687	0.003821624	red
-0.248711666	0.006936259	red
0.000849715	0.00154234	red
-0.00022143	0.002328499	red
-0.003455077	0.004070711	red
-0.497544241	0.562919366	red
-0.50948386	0.020441263	red
-0.505997858	0.024710863	red
-0.250547132	0.182944448	red
-0.250533325	0.004608016	red
-0.245102934	0.009676318	red
-0.002839939	0.000826124	red
-0.002986054	0.002014208	red
-5.69E-05	0.085308108	red
-0.502362333	0.620513473	red
-0.501894802	0.012849757	red
-0.496545604	0.577934621	red
-0.252606681	0.183038182	red
-0.250019362	0.176398564	red
-0.257351069	0.00832459	red
0.000606852	0.000791066	red
0.000162118	0.001866498	red
-0.000560199	0.004047847	red
-0.496006069	0.615688806	red
-0.50034425	0.603622354	red
-0.503083358	0.025208325	red
-0.250777722	0.001279469	red
-0.256550042	0.17506188	red
-0.248630027	0.170698278	red
-0.005298025	0.000513028	red
0.001485413	0.001972453	red
-0.00055119	0.00408833	red
-0.500824041	0.003705615	red
-0.502023989	0.016444885	red
-0.5031889	0.025163581	red
-0.245384745	0.000636265	red
-0.242572199	0.004190602	red
-0.252818456	0.006471755	red
0.003626604	0.000492419	red
0.000650903	0.002864399	red
-0.00506812	0.003962159	red
-0.506141007	0.002351026	red
-0.502405246	0.019407177	red
-0.507672688	0.028940176	red
-0.250454446	0	red
-0.249957023	0.168897281	red
-0.256242184	0.007673984	red
-0.003038866	0.000443767	red
0.002928592	0.002426168	red
-0.003273789	0.085695381	red
-0.505361745	0.20370463	red
-0.501396703	0.200781345	red
-0.504907978	0.199538788	red
-0.249368827	0.060110866	red
-0.255336272	0.05904324	red
-0.250230327	0.058833074	red
0.007270642	0.029332344	red
-0.004937501	0.02978083	red
-0.002669822	0.030030568	red
-0.494533605	0.203666701	red
-0.513449747	0.204900988	red
-0.500627582	0.19913926	red
-0.244235323	0.059036788	red
-0.256077132	0.059992775	red
-0.248662736	0.059411245	red
0.002280567	0.02897972	red
0.006367937	0.02994052	red
0.000179086	0.030522389	red
-0.498853727	0.198137307	red
-0.503784419	0.200999514	red
-0.495762594	0.202094272	red
-0.251177126	0.059319183	red
-0.257742049	0.061329636	red
-0.250919139	0.059908451	red
-0.004794984	0.029741448	red
0.000166006	0.029577351	red
0.006347658	0.029171884	red
-0.503646319	0.199929753	red
-0.503471376	0.197697719	red
-0.503198151	0.20079284	red
-0.252653929	0.059519195	red
-0.249199586	0.059097894	red
-0.252953031	0.059728899	red
-0.000605835	0.029930464	red
0.001281734	0.029252957	red
-3.63E-05	0.029318015	red
-0.505985835	0.201805711	red
-0.491833578	0.203506184	red
-0.490921079	0.201879001	red
-0.252317722	0.058818804	red
-0.248205021	0.060335486	red
-0.249806103	0.059759755	red
0.001893501	0.029054078	red
-0.001937429	0.029675657	red
1.92E-05	0.029908405	red
-0.501908592	0.196582919	red
-0.497810851	0.199909072	red
-0.496622791	0.198324002	red
-0.25856749	0.058481836	red
-0.249287784	0.059107116	red
-0.248443619	0.059432318	red
-0.00200393	0.03075595	red
-0.000657757	0.029826697	red
-0.000797129	0.030094754	red
-0.504211168	0.203645555	red
-0.494980248	0.204340816	red
-0.505778569	0.20193633	red
-0.257269869	0.059214376	red
-0.250528621	0.059867531	red
-0.245990811	0.059648256	red
0.004492417	0.029627915	red
7.31E-05	0.029202207	red
0.001659014	0.029584023	red
-0.50366966	0.199963453	red
-0.500767389	0.199089324	red
-0.504113429	0.198577955	red
-0.24544868	0.060652757	red
-0.257647492	0.059553779	red
-0.248902526	0.059034628	red
0.004160312	0.029793733	red
0.007688275	0.029434676	red
0.000927052	0.029681951	red
-0.497899356	0.199087195	red
-0.497632728	0.198595318	red
-0.510251428	0.204410409	red
-0.249210842	0.059716422	red
-0.250060277	0.059222005	red
-0.247572927	0.058777889	red
-0.006018339	0.029836326	red
-0.004147481	0.029660495	red
0.006398344	0.029694606	red
-0.499120296	0.134806012	blue
-0.498286944	0.193038877	blue
-0.506839901	0.043897452	blue
-0.25610811	0.04032638	blue
-0.248779399	0.05677324	blue
-0.25400125	0.011554791	blue
-0.00276232	0.019561428	blue
-0.000527651	0.02880921	blue
0.005069197	0.006421057	blue
-0.507136465	0.130467929	blue
-0.494725146	0.186909534	blue
-0.492440318	0.036357469	blue
-0.254877717	0.039560755	blue
-0.250922637	0.054835119	blue
-0.244325831	0.010162578	blue
0.003975366	0.019358325	blue
-0.003125187	0.027310823	blue
0.000312368	0.004881079	blue
-0.494055464	0.130916368	blue
-0.502407076	0.178891371	blue
-0.504773938	0.029647865	blue
-0.248437181	0.036996665	blue
-0.252215259	0.053137874	blue
-0.248512669	0.008893327	blue
0.010341967	0.019132542	blue
0.002831293	0.026244597	blue
-0.008423304	0.004394364	blue
-0.500766677	0.136420355	blue
-0.505079501	0.191866629	blue
-0.490996117	0.042192534	blue
-0.244580454	0.040054978	blue
-0.248301423	0.057051882	blue
-0.252667671	0.013005116	blue
-0.00311538	0.019845286	blue
-0.002290955	0.028917744	blue
-0.000420892	0.006509741	blue
-0.500878422	0.131686755	blue
-0.511900264	0.185650881	blue
-0.505833186	0.036922346	blue
-0.250962858	0.039834416	blue
-0.247185727	0.054079953	blue
-0.244945881	0.010504285	blue
-0.002159412	0.019662536	blue
0.007133085	0.027613882	blue
0.000155882	0.005433296	blue
-0.496803354	0.133437565	blue
-0.501892651	0.175741316	blue
-0.496627453	0.031300633	blue
-0.243998229	0.038214722	blue
-0.251300798	0.051334806	blue
-0.247430535	0.007692148	blue
0.001015738	0.019339976	blue
-0.010221912	0.026392668	blue
0.000937875	0.00355601	blue
-0.490679794	0.136368342	blue
-0.497413626	0.193057912	blue
-0.496752629	0.044781319	blue
-0.261234871	0.038953207	blue
-0.254119039	0.056538694	blue
-0.248417676	0.013501032	blue
-0.004111808	0.019578292	blue
-0.003078037	0.028461038	blue
0.00041969	0.00677071	blue
-0.501217164	0.136941884	blue
-0.497039866	0.185894163	blue
-0.495532211	0.037282556	blue
-0.25470211	0.039356411	blue
-0.247255941	0.054675587	blue
-0.253227149	0.009903217	blue
0.00180488	0.019545363	blue
-0.006414988	0.027756904	blue
0.006973035	0.00567221	blue
-0.493888434	0.133893843	blue
-0.50257358	0.181231433	blue
-0.490169763	0.008584772	blue
-0.245104856	0.038018317	blue
-0.249933903	0.052706875	blue
-0.2534709	0.009934853	blue
-0.011642485	0.019352254	blue
-0.002890144	0.026800424	blue
-0.002096569	0.004548313	blue
-0.251828633	0.00822198	green
-0.246530215	0.010744013	green
-0.00306562	0.003433657	green
0.004564098	0.004601146	green
-0.002784992	0.005374734	green
-0.497740159	0.025852707	green
-0.509319578	0.029633319	green
-0.494008294	0.035735396	green
-0.24745628	0.005761677	green
-0.25782003	0.008909478	green
-0.245424111	0.009172183	green
-0.001536154	0.002852014	green
0.001886403	0.004486226	green
-0.001354893	0.005320014	green
-0.494073981	0.026577184	green
-0.499589783	0.029059811	green
-0.504019892	0.041427634	green
-0.254372254	0.007460489	green
-0.247863049	0.009234736	green
-0.255148621	0.011716473	green
0.001518478	0.00345072	green
-0.002013545	0.004328549	green
-0.004365001	0.006200761	green
-0.504191075	0.022852484	green
-0.495880796	0.037070346	green
-0.50955763	0.03550692	green
-0.257095067	0.006778215	green
-0.253429439	0.010263421	green
-0.256676428	0.011012706	green
-0.009626353	0.003711149	green
-0.000927499	0.004110648	green
-0.004966871	0.005084316	green
-0.49562384	0.019366463	green
-0.500706534	0.026936203	green
-0.505183989	0.037044948	green
-0.249599553	0.007651823	green
-0.264055178	0.008172085	green
-0.247925335	0.00961765	green
-0.006966483	0.002777466	green
-0.011464459	0.004318773	green
-0.004848662	0.005281639	green
-0.500782799	0.026046729	green
-0.498247804	0.033563619	green
-0.49645277	0.041216064	green
-0.250616723	0.008376902	green
-0.248393406	0.009108658	green
-0.255279362	0.011644838	green
-0.000236264	0.002898456	green
0.000579694	0.004514825	green
0.008820777	0.004789463	green
-0.497013234	0.023862921	green
-0.504208404	0.028389991	green
-0.505542966	0.036434724	green
-0.257271988	0.005119028	green
-0.254041598	0.009313388	green
-0.250728083	0.4	green
-0.005828252	0.003836292	green
-0.010860367	0.004813058	green
-0.009832646	0.005461129	green
-0.500522974	0.028184597	green
-0.508997077	0.034476716	green
-0.502799499	0.037156814	green
-0.25293277	0.00741234	green
-0.250311021	0.008178172	green
-0.24475962	0.011048544	green
-0.000272614	0.003645023	green
-0.001104602	0.004000092	green
-0.000769198	0.005487728	green
-0.498976382	0.029017352	green
-0.505696881	0.030767139	green
-0.512307783	0.036307705	green
-0.246570056	0.007859787	green
-0.256855357	0.008197739	green
-0.254212128	0.010164406	green
-0.004233296	0.003158391	green
-0.000157778	0.004835252	green
0.006041485	0.005437431	green
-0.497419751	0.022624441	green
-0.503932021	0.026227852	green
-0.493979817	0.038176459	green
-0.24749843	0.005717543	green
};

\legend{Small, Medium, Large}
\end{axis}

\node[draw = none,text width=4cm] at (6.25,-1.5) {\textbf{Installed base size}};

\end{tikzpicture}

    \caption{Fraction of AM in the installed base for different levels of the AM failure rate relative to the CM failure rate, prescribed by the DCL policy.}
    \label{fig:scatterplot}
\end{figure}
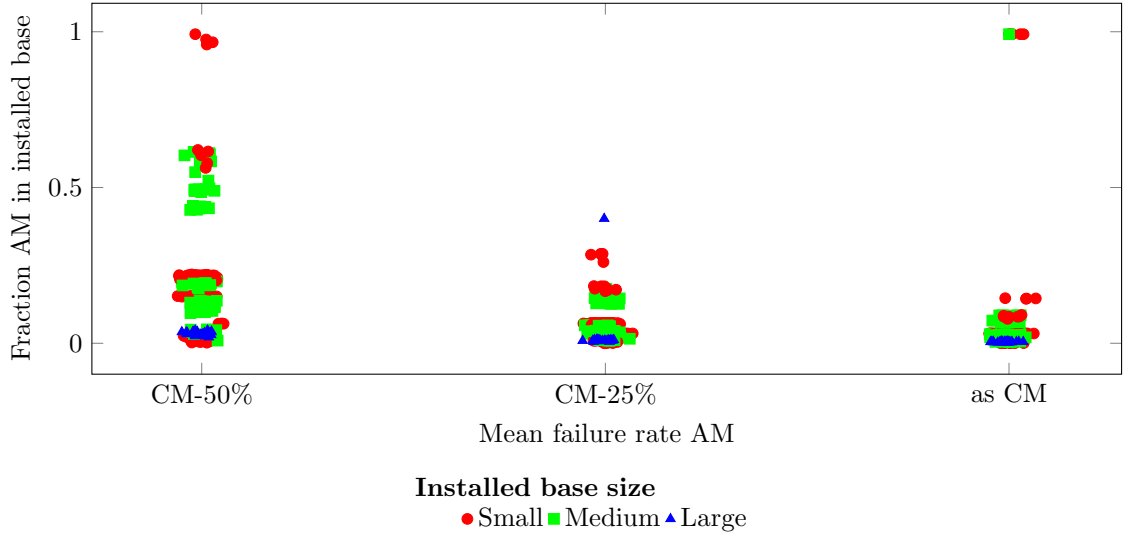

\newpage

{In Figure~\ref{fig:endo_study}, we analyse the impact of ignoring endogenous demand. Specifically, we remove the assumption that demand is endogenous and instead assume equal failure rates for AM and CM, i.e., $\mu_C = \mu_A$. First, we train \gls{DCL} under this standard setting, considering only Item 1. Next, we train \gls{DCL} while accounting for endogenous demand, allowing $\mu_A$ to vary. In the figure, the x-axis represents the $\mu_A/\mu_C$ ratio, while the y-axis shows the percentage increase in costs for the \gls{DCL} policy that ignores endogenous demand. We verified that dual sourcing is profitable across all settings, outperforming single sourcing from either AM or CM. Our findings align with the insights of \citet{PERON2022}: (i) ignoring endogenous demand can result in significant cost increases, with observed increases of up to $9\%$, (ii) when the failure rates of parts are similar ($0.8 \leq \mu_A/\mu_C \leq 1.1$), the cost impact of disregarding endogenous demand is minimal, and
(iii) when the AM failure rate is lower than the CM failure rate, the costs of neglecting endogenous demand rise more rapidly compared to when the opposite is true. Note that the absolute costs for all policies are higher when $\mu_A/\mu_C>1$, as more failures occur on average and a higher stock is needed. We note that this is an empirical observation and it may not generalise to other dual sourcing cases.}

{\begin{observation}
    Ignoring demand to be endogenous can result in a significant increase in costs, especially when the expedited part has a lower failure rate compared to the regular part.
\end{observation}}

\begin{figure}[hbtp]
    \centering
    \begin{tikzpicture}[scale = 0.84]
            \begin{axis}[
                legend style={at={(0.5,-0.2)},anchor=north, draw=none},
                legend columns=2,
                %symbolic x coords={1,2,3,4},
                xtick distance=1/5,
                ymin = 0 ,ymax=15,
                xmin=0.09, xmax=2.01,
                 xlabel = AM/CM failure rate ratio $\mu_A/\mu_C$,
                xlabel near ticks,
                ylabel= Costs increase compared to retrained DCL,
                ylabel near ticks,
                legend cell align={left},
                 yticklabel={$\pgfmathprintnumber{\tick}\%$},
                        ]

        \addplot[ mark options={fill=black, mark size=2.5pt},mark=diamond*] %
                coordinates {
                    (0.1,9)
                    (0.2,8)
                    (0.3,7.3)
                    (0.4,7)
                    (0.5,4)
                    (0.6,3)
                    (0.7,1.2)
                    (0.8,0.8)
                    (0.9,0.29)
                    (1.0,0)
                    (1.1,0.3)
                    (1.2,1.6)
                    (1.3,2.1)
                    (1.4,2.5)
                    (1.5,3)
                    (1.6,5)
                    (1.7,6.6)
                    (1.8,7.2)
                    (1.9,7.5)
                    (2.0,7.9)

                };

            \end{axis}
            \node at (3.5,6.0) {Effect of supply-mode dependent demand};
    \end{tikzpicture}
    \caption{Average performance of \gls{DCL} trained on an instance where demand is non-endogenous ($\mu_C=\mu_A$) and applied to instances where demand is endogenous ($\mu_A$ is varied), compared to DCL which is retrained for each specific setting, results over the standard Item~1 instance over $100$ replications.}\label{fig:endo_study}
\end{figure}

Figure~\ref{fig:stacked_bars} illustrates key differences between DCL and IWA in their ordering behaviour. First, \gls{DCL} has more flexibility, allowing it to order any AM quantity, leading to more frequent large-quantity orders, whereas \gls{IWA} follows fixed reorder points and does one-for-one replenishment. Second, when \gls{DCL} orders AM, it seems to order larger batches. It seems that this is caused by the shorter lead time and lower failure rates of \gls{AM} for many instances, see Table~\ref{table:oilgasparameters}. This aligns with the findings of \citet{Westerweel2020PrintingManufacturing}, which indicate that in remote areas, AM can be the preferred source due to shorter lead times. Overall, we observe that $41.8\%$ of the IWA orders are AM, whereas \gls{DCL} orders $71.6\%$ AM.

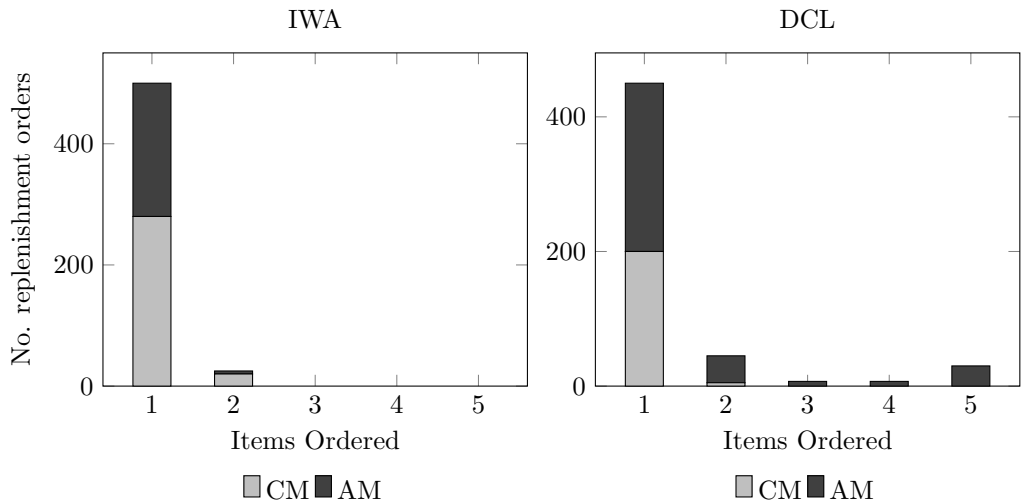
\begin{figure}[hbtp]
    \centering
    \begin{tikzpicture}
        \begin{axis}[
            width=0.45\textwidth,
            height=6cm,
            bar width=0.5cm,
            ybar stacked,
            symbolic x coords={1, 2, 3, 4, 5},
            xtick=data,
            ylabel={No. replenishment orders},
            xlabel={Items Ordered},
            title={IWA},
            ymin=0,
            enlarge x limits=0.15,
            every axis plot/.append style={draw=black, fill=black},
            legend style={draw=none,at={(0.5,-0.25)},anchor=north,legend columns=-1},
        ]
            \addplot+[
                color=black,fill=black!25
            ] coordinates {(1,280) (2,20) (3,0) (4,0) (5,0)};
            \addplot+[
                color=black,fill=black!75
            ] coordinates {(1,220) (2,5) (3,0) (4,0) (5,0)};
            \legend{CM, AM}
        \end{axis}
    \end{tikzpicture}
    \begin{tikzpicture}
        \begin{axis}[
            width=0.45\textwidth,
            height=6cm,
            bar width=0.5cm,
            ybar stacked,
            symbolic x coords={1, 2, 3, 4, 5},
            xtick=data,
            xlabel={Items Ordered},
            title={\gls{DCL}},
            ymin=0,
            enlarge x limits=0.15,
            every axis plot/.append style={draw=black, fill=black},
            legend style={draw=none,at={(0.5,-0.25)},anchor=north,legend columns=-1},
        ]
            \addplot+[
                color=black,fill=black!25
            ] coordinates {(1,200) (2,5) (3,0) (4,0) (5,0)};
            \addplot+[
                color=black,fill=black!75
            ] coordinates {(1,250) (2,40) (3,7) (4,7) (5,30)};
            \legend{CM, AM}
        \end{axis}
    \end{tikzpicture}
    \caption{Observed orders for all non-dominated instances for the \gls{IWA}(left) and \gls{DCL} (right) policies}
    \label{fig:stacked_bars}
\end{figure}

\begin{observation}
    {A dual sourcing policy can save costs} by ordering larger quantities and more often relying on the AM source.
\end{observation}

As a final analysis, we investigate which problem parameters explain the relative performance of \gls{IWA} compared to \gls{DCL}. This way, we can provide clear guidelines for the instances in which IWA is preferred over DCL. To achieve this, we conduct a Mann-Whitney test to identify significant correlations with the relative savings, we find that the installed base size, lead time of \gls{CM} and \gls{AM}, holding costs, and backorder costs show significant correlation to the relative savings with P-values of $0.014$ and less. Using these parameters, we fit a logistic regression model to predict whether \gls{IWA} or \gls{DCL} should be used for a specific problem instance. The accuracy of this logistic regression model (defined as $\frac{\textrm{Number of correct predictions}}{\textrm{Total number of predictions}}$) is $0.892$.

Analyzing the model coefficients, we find that DCL is preferred in most cases, but whenever (i) the installed base is small, (ii) the backorder costs are high, (iii) the lead time of AM and/or CM is high, or (iv) the holding costs are high, it could be valuable to evaluate if IWA outperforms DCL. \gls{DCL} is generally more suitable for instances with larger installed bases or higher backorder costs, where the flexibility of \gls{DCL} is better suited to the increased demand. We conjecture that \gls{DCL}'s flexibility allows for dynamic adjustment to varying conditions, but this same flexibility can lead to overfitting and variability in certain high-cost or high-lead-time environments. On the other hand, \gls{IWA} is more appropriate for scenarios with longer lead times or higher holding costs, where its efficiency and simplicity provide a better balance between inventory levels and replenishment schedules. This differentiation can guide practitioners in selecting the most effective policy based on specific problem settings. 
\begin{observation}
    While DCL is generally preferred, IWA may be more effective for instances with small installed bases, high backorder costs, long lead times, or high holding costs.
\end{observation}

\section{Conclusion}
\label{sec:Conclusion}

We addressed the dual sourcing problem with supply mode dependent demand, developing an intuitive heuristic approach and leveraging data-driven methods. We compare three distinct classes of reinforcement learning (RL) policies, i.e., value-based, policy-based, and actor-critic. So far, the supply mode dependent demand setting was not studied in a realistic setting before. Our paper is the first to study realistic problem sizes, beyond the capabilities of analytical approaches. In our numerical study, we first studied small instances to benchmark against an exact approach, after which we studied a real-world case from the energy sector. 

The best performing methods are iterative weight adjustment (IWA), which finds a policy in seconds, and the Deep controlled learning (DCL) RL approach, with an average optimality gap of $1.9\%$ and $0.4\%$, respectively. On the real-world energy case these policies also perform well, saving $16.9\%$ and $22.6\%$ in total costs compared to single sourcing, respectively. Furthermore, we present an approach called Endogenously parameterised learning (EPL), which extends learning-based methods, such as DCL, and allows them to be trained once and next be applied to a variety of SKUs. DCL embedded into EPL finds an average optimality gap of $0.4\%$ and in the real-world case performs only $3\%$ worse than DCL, at a fraction of the required computational time ($52\times$ less). Our observations lead to the following synthesis of insights:
\begin{itemize}
    \item Learning-based dual sourcing methods like DCL are less effective when the performance difference between optimal dual sourcing policies and single sourcing heuristics is small.
    \item Training on multiple SKUs simultaneously enhances policy robustness and efficiency.
    \item DCL significantly reduces costs compared to both the IWA heuristic and a single sourcing benchmark.
    \item EPL offers competitive performance with reduced computational effort.
    \item DCL's cost savings stem from larger, more frequent orders from the AM source.
    \item IWA may be better suited for scenarios with small installed bases, high backorder or holding costs, and long lead times.
    \item {Ignoring demand to be endogenous may increase costs significantly, depending on the difference between the AM and CM failure rate.}
    \item Dual sourcing is in general advantageous for small to medium sized installed bases with a low AM failure rate compared to the CM failure rate.
\end{itemize}
While the simple IWA heuristic is generally sufficient to obtain near-optimal solutions and offers substantial computational speed, DCL often outperforms IWA in specific instances, particularly by reducing backorder costs. The EPL approach shows promise by providing significant computational savings and robust policy performance across diverse problem settings.

Future research can focus on several directions. Investigating more advanced sampling procedures for the EPL approach, such as those from the optimal learning field (cf. \citet{PowellBookOptimal}), can improve the efficiency and accuracy of parameter sampling. EPL could be used for pre-training on a variety of SKUs, followed by transfer learning to fine-tune on specific or new SKUs. Applying the EPL approach to other inventory control problems could further demonstrate its potential across various contexts and industries. Finally, incorporating stochastic lead times, different demand distributions, non-stationary demand caused by additional information resulting from predictive maintenance, and multi-echelon systems would be logical extensions of our problem. 

\section*{Acknowledgements}
This research has partially been conducted within the project \enquote*{Sustainability Impact of New Technology on After sales Service supply chains}  (SINTAS) that has been sponsored by the Netherlands Organisation for Scientific Research under Project Number 438-13-2.

Fabian Akkerman conducted his research in the project DynaPlex: Deep Reinforcement Learning for
Data-Driven Logistics, made possible by TKI Dinalog and the Topsector Logistics and funded by
the Ministry of Economic Affairs and Climate Policy (EZK) of the Netherlands.

%prints the bibliography of the used citations
\bibliographystyle{elsarticle-harv}
\bibliography{references}
\appendix

\section{{Supplementary Results}}\label{app:extra_results}

{In Table~\ref{tab:results_app}, we show the results of two additional settings of the synthetic case. For these two instances, we use the settings for Instances 1 and 2, respectively. However, we increase the installed base size from $7$ to $20$. Due to the increase of the state space, the exact policy becomes intractable, so we no longer show the percentage gap to the optimal policy, but show the percentage improvement over the BSP policy. A negative sign indicates that the policy performs worse than the BSP policy. We observe that, similarly to the results in Table~\ref{tab:results}, IWA, DCL, and \gls{DCL}$^\textrm{EPL}$ are the best performing policies. However, with this larger instance size, the performance of AVI and AVI$^\textrm{EPL}$ is considerably worse. \gls{PPO} and \gls{PPO}$^\textrm{EPL}$ show similar results as before and are unable to outperform the BSP policy.}

\begin{table}[hbtp]
    \caption{Percentage improvement over the BSP policy, the best performing policies are \textbf{highlighted}.}
    \label{tab:results_app}
    \centering
    \begin{tabular}{l  c c c c c c c c }
    \toprule
    Instance & \gls{IWA}& \gls{AVI} & AVI$^\textrm{EPL}$ & \gls{DCL} & \gls{DCL}$^\textrm{EPL}$ & \gls{PPO} & \gls{PPO}$^\textrm{EPL}$ \\
    \midrule 
 11   & \textbf{6.7\% } & -0.8\% & 0\%  & 5.8\%  & 2.5\%  & -1.7\%  & -0.8\%  \\
    12 & 6.8\%  & 0\%  & 0\%  & \textbf{7\% } & 4.8\%  & -0.3\%  &  0\%  \\
    \bottomrule
    \end{tabular}
\end{table}

\section{Transition Function}\label{app:transitions}
We detail the calculation of the values for $y_{C,t}, y_{A,t},z_{C,t}$, and $z_{A,t}$, which are used as helper variables to support the transition from a state $\mathbf{i}_t$ to $\mathbf{i}_{t+1}$.
\begin{itemize}
\item $y_{C,t}=min\{B_{t}+k_{C,t}+k_{A,t},s_{C,t}\}$ where the first element describes the total demand (backorders + new demand arrivals in period $t$), and the second element the \gls{CM} on-hand stock. 
\item $y_{A,t}=min\{B_{t}+k_{C,t}+k_{A,t}-y_{C,t},s_{A,t}\}$ where the first element describes the total demand not satisfied by \gls{CM} before a possible order arrival, and the second element the \gls{AM} on-hand stock.
\item $z_{C,t}=min\{B_{t}+k_{C,t}+k_{A,t}-y_{C,t}-y_{A,t},A_{C,t}\}$ where the first element describes the backlog just before a possible order arrives, and the second $A_{C,t}$ element is equal to the number of \gls{CM} units that may arrive at $t$. 
\item $z_{A,t}=min\{B_{t}+k_{C,t}+k_{A,t}-y_{C,t}-y_{A,t}-z_{C,t},A_{A,t}\}$ where the first element describes the remaining backlog after we accounted for the \gls{CM} order arrival, and the second element $A_{A,t}$ is equal to the size of a potential \gls{AM} order arrival.
\end{itemize}

\section{Exact Model and Solution}
\label{app:simpModel}
We elaborate how the problem described in Section~\ref{sec:prob} may be simplified to obtain a model in which we do not distinguish between \gls{CM} and \gls{AM} items.
\medbreak
Any combination of $n_{C,t}$ and $n_{A, t}$ is replaced by $n_{t}$. The same holds for  $s_{C,t}$ , $s_{A, t}$ and thus we define $\mathbf{i}_{t}=(IP_t,n_{t},s_{t},u_{C,t,t^\prime},u_{A,t,t^\prime})$. The decision space and cost calculations remain unchanged, except that we do not differentiate between \gls{AM} and \gls{CM} failures any longer. Hence, the expected demand in period $t$ is equal to the \enquote*{general} failure rate ($\mu$) multiplied with the number of operating parts ($n_{t}$), and we have to change the calculation of the maintenance and backorder costs accordingly. The number of operating systems at the beginning of period $t+1$ simplifies to:
\begin{equation}
n_{t+1}= n_{t}-k+x,
\end{equation}
where $x$ is equal to the number of items that may be installed \textit{before} and \textit{after} a possible replenishment order arrives in period $t$, i.e., $x=\min\{B_{t}+k,s_{t}+A_{t}\}$, where the first element describes the total demand (backorders + new demand arrival in period $t$), and the second element the on-hand stock plus the total number of items $A_{t}$ arriving at $t+\tau$.
Finally, we simplify the computation of the on-hand stock in period $t+1$ to:
\begin{equation}
s_{t+1}= s_{t}+A_{t}-x.
\end{equation}

To measure the weight $\rho$ of sourced \gls{AM} items, we multiply the steady state probabilities with the corresponding order quantities.

% \section{Exact Solution Procedure}
% \label{app:SolProMDP}
In Algorithm~\ref{alg:Policy iteration}, we describe how the policy iteration algorithm may be used to find the optimal policy $\pi^s$ for the Bellman Equations~\ref{eq:bellEQ}.
\begin{algorithm}[h]
Set $j = 1$ and chose a stationary policy $\pi_{j}$\;
\Repeat{$\pi_{j}=\pi_{j-1}$}{
    Determine $g(\pi_{j}), v(\mathbf{i}_{t},\pi_{j})$ with the system of linear equations\;
$v(\mathbf{i}_{t},\pi_{j})=C(\mathbf{i}_{t},\pi_{j})-g(\pi_{j})+\sum_{\mathbf{i}_{t+1}\in\mathcal{I}_{t+1}} p(\mathbf{i}_{t+1}|\mathbf{i}_{t},\pi_{j})v(\mathbf{i}_{t+1},\pi_{j}), \forall \mathbf{i}_{t}\in\mathcal{I}$\;
$v(\mathbf{s}_{t},\pi_{j})=0$ with $\mathbf{s}_{t}$ an arbitrary state\;
Find the new policy $\pi_{j+1}$\;
$(x_C, x_A)_{j}=argmin_{(x_C, x_A)\in\mathcal{A}}\{C(\mathbf{i}_{t},\pi_{j})-g(\pi_{j})+\sum_{\mathbf{i}_{t+1}\in\mathcal{I}_{t+1}} p(\mathbf{i}_{t+1}|\mathbf{i}_{t}, \pi_{j})v(\mathbf{i}_{t+1},\pi_{j})\}$\;
$\pi_{j+1}(\mathbf{i}_{t}) = (x_C, x_A)_{j}, \forall \mathbf{i}_{t}\in\mathcal{I}$\;
Set $j=j+1$\;
}
\Return the optimal policy $\pi_{j}$
  	\caption{Policy iteration}
	\label{alg:Policy iteration}
\end{algorithm}
% \newpage
\section{IWA}\label{app:ira}

In Algorithm~\ref{alg:Iterative procedure}, we detail the algorithmic steps for obtaining the IWA heuristic policy. In our experiments, we set $\psi=0.2$.
\begin{algorithm}[hbtp]
	Set $j:=1$, $\gamma_{1}:=0.0$ and $\psi$\;
    \Repeat{$|\gamma_{j-1}-\gamma_{j}|<\psi$}{
    Compute $\mathbb{E}[X]$ and $Var[X]$ with Equation~\ref{6eq:expect} and \ref{6eq:var}\;
    Measure $\rho_{j}$ based on $\pi^{S}(\mathbb{E}[X],Var[X])$\;
    Compute $\gamma_{j+1}$ using $\rho_{j}$ and Equation~\ref{6eq:Gamma}\;
    $j=j+1$\;}
   \Return{Heuristic policy}
  	\caption{Iterative weight adjustment}
	\label{alg:Iterative procedure}
\end{algorithm}

\section{Dual Index Policy}\label{sect:dualindex}
For small instances, we can calculate the optimal single demand rate for the policy $\pi^S$, as shown in Appendix~\ref{app:simpModel}. However, for realistic instances, obtaining an optimal $\pi^s$ is no longer tractable. Hence, we employ the dual index policy to find a reordering policy used in IWA. The optimal dual index policy closely replicates the decisions of the optimal state-dependent policy derived through dynamic programming, thereby achieving near-optimal performance in the majority of scenarios \citep{li2014,hua2015,tang2023}. The dual index policy is defined by two parameters $z_C$ and $z_A$, where $z_C$ is the order-up-to level for \gls{CM} parts and $z_A$ is the order-up-to level for \gls{AM} parts. At the start of a period, orders are placed based on the \gls{AM} and \gls{CM} inventory positions, respectively. To support this, we add separate inventory positions to our state definition. The \gls{AM} order is added to the \gls{CM} inventory position before the \gls{CM} order is determined. Compared to a standard base-stock policy, the dual index policy may \emph{overshoot} the targeted inventory position $z_A$ of \gls{AM} parts. This overshoot is caused by the new information of the order $x_C$ placed $l=l_C-l_A$ periods ago, see Equation~\ref{eq:invenP_CM} and Equation~\ref{eq:invenP_AM}. 
\begin{align}
IP_{C,t}&=s_{C,t}+(u_{A,t,t-l_A}+\ldots+u_{A,t,t-1})+(u_{C,t,t-l_C}+\ldots+u_{C,t,t-1}), \label{eq:invenP_CM}\\
IP_{A,t}&=s_{C,t,t}+(u_{A,t,t-l_A}+\ldots+u_{A,t,t-1})+(u_{C,t,t-l_C}+\ldots+u_{C,t,t-l-1}).\label{eq:invenP_AM}
\end{align}

This new information may cause the inventory position of \gls{AM} parts to exceed its target $z_A$. We define $\Delta = z_C-z_A$. In \citet{Veeraraghavan2008NowSystems}, it is shown that the overshoot distribution is a function of $\Delta$ independent of $z_A$. For each $\Delta$, an expression can be derived for the optimal $z_A^*(\Delta)$, as a newsvendor fracticle of the demand during lead time convoluted with the overshoot. By a one-dimensional search over $\Delta$, we can find the lowest costs ($\Delta$, $z^*_A(\Delta)$) pair. This yields the optimal dual index policy. For complete details and derivations of the dual index policy, we refer to \citet{Veeraraghavan2008NowSystems}.

\section{\gls{AVI} }\label{subsec:ADP}
Algorithm~\ref{alg:VIfp} presents a high-level summary of \gls{AVI}. This version implements a direct update through a forward pass coupled with pure exploration \citep{powellBookNew}. We represent a value function approximation by $\overline{V}$, a sequence of randomly sampled demand (sample path) by $\omega \in \Omega$, and the downstream costs by $\Hat{v}$. It is important to note that the approximated downstream costs are encapsulated within $\Hat{v}$, employing a bootstrap estimate \citep{sutton}. \gls{AVI} necessitates frequent checks for convergence since the derived policy $\widehat{\pi}$, that is, selecting the decision with the lowest expected discounted costs in a greedy manner, may have stabilised while the value function $\overline{V}$ has not.

After initialisation of the approximation $\overline{V}_0$ to zero in Step 0, $N$ samples are collected by following a trajectory of demand $\omega$ (sample path)  of length $N$, starting in initial state $s_0$.  During one iteration, first a simulation step is conducted by finding the decision that returns the lowest sum of the direct and estimated downstream costs, i.e., following the last updated policy $\widehat{\pi}_{n-1}$ based on $\overline{V}_{n-1}$ that is supplied with the results from transition function $S^M$, in Step 2a. The subsequent downstream costs $\Hat{v}_{n-1}$ are stored in Step  2a, except if the algorithm is still in the first iteration ($n>0$).

Next, a new post-decision state is sampled. There are different options for sampling new states, e.g., incorporating a random (not necessarily feasible) decision or incorporating a policy decision, or drawing new states. For this paper, we use $\epsilon$-greedy exploration method, i.e., we use random decisions $\widetilde{a}_n$ with probability $\epsilon$ to find new post-decision states, and otherwise follow the current policy in Step 2b. 

The feature values ${\phi}_{n-1}$ related to the subsequent post-decision state $S_{n-1}^{\widetilde{a}}$ are stored in Step 2c and the move to the next state is made using sampled event $\omega_n$ in Step 2d. If we are not in the first iteration ($n>0$), the previous value function $\overline{V}_{n-1}$ is updated directly using a single data point, i.e., using $\phi_{n-1}$ and $\Hat{v}_{n-1}$ we obtain $\overline{V}_{n}$ in Step 2e. 

Since the updates of the value function are conducted directly, for the new state $S_{n+1}$ a new and improved value function $\overline{V}_{n}$ can directly be used for the next simulation step.

\begin{algorithm}[hbtp]\caption{\gls{AVI}}\label{alg:VIfp}

\scriptsize

        Step 0. \qquad \textbf{Initialisation} \\
        
            \hskip5.5em Step 0a. \qquad Initialise the approximation for $\overline{V}_0$ \\
            
             \hskip5.5em Step 0b. \qquad Set the iteration counter to $n=0$, and number of iterations $N$ \\
             
        Step 1. \hskip2em Choose a sample path $\omega \in \Omega$ and initial state $s_0$ \\
            
        Step 2. \hskip2em  \textbf{Policy Extraction and Value Function Updating}\\
        
        \hskip5.5em Step 2a. \hskip1.8em If $n>0$, find feasible decision $a_{n}$ that solves \\ 
		  \hskip9.5em \qquad$  \Hat{v}_{n-1} = \min\limits_{a\in\mathcal{A}(s_n)} C(s_n,a) + \gamma \overline{V}_{n-1}(s^a_n)$\\
        
        \hskip5.5em Step 2b. \hskip1.8em Transition to post-decision state $s_{n}^{\widetilde{a}}$ with random or policy decision $\widetilde{a}_n$  \\
        
        \hskip5.5em Step 2c. \hskip1.9em Store feature values ${\phi}_{n}$ of state $s_{n}^{\widetilde{a}}$ \\
        
         \hskip5.5em Step 2d. \hskip1.8em Transition to state $s_{n+1}$ with sampled event $\omega_{n}$ \\
        
		\hskip5.5em Step 2e. \hskip1.95em If $n>0$, \textbf{update approximation} $\overline{V}_{n} \leftarrow \overline{V}_{n-1}$ with ${\phi}_{n-1}$ and $\Hat{v}_{n-1}$\\

		Step 3. \qquad {Increment $n$, if $n\leq N$ go to Step 2, else return $\overline{V}_N$}

\end{algorithm}

We employ linear regression in AVI. Linear regression requires some feature engineering. We provide the following state features to the \gls{AVI} algorithm: each state variable, and the following other features:
\medbreak
\begin{itemize}
\item $IP_{t}$: the inventory position of \gls{CM} and \gls{AM} parts as defined by Equation~\ref{eq:ip}.
\item $\mathbb{E}[D_{t}]$: the expected demand in period $t$, i.e., $\mathbb{E}[D_{t}]=n_{C, t}\mu_{C}+n_{A,t}\mu_{A}$.
\item $Var[D_{t}]$: the demand variance in period $t$, i.e., $Var[D_{t}]=n_{C, t}V_{C}+n_{A,t}V_{A}$.
\item $IL_{t}$: the inventory level in period $t$, i.e., $IL_{t}=s_{C,t}+s_{A,t}-B_{t}$.
\end{itemize} 
\medbreak
The $IP_t$ feature is commonly used in dual sourcing heuristics. The features $\mathbb{E}[D_{t}]$ and $Var[D_{t}]$ provide insight into the state dependent failure behaviour, and $IL_{t}$ gives an indication of the stock-out risk/severity. When embedding \gls{AVI} in EPL, we also provide the problem parameters as additional features, see Section~\ref{sect:paramEPL}.

\section{{\gls{DCL}}}\label{subsec:DCL}

\gls{DCL} as proposed in \citet{temizoz2023deep} is a \gls{RL} framework based on approximate policy iteration. The \gls{DCL} algorithm is designed for problems that deal with high-stochasticity and high variance in the cost incurred over a horizon. Hence, it is suitable for our problem as only a small part of the state space requires \gls{AM} ordering, which causes high variability in costs. \gls{DCL} estimates the optimal decisions for each state in a data set by revisiting and evaluating that state under multiple exogenous scenarios and selecting the decision with the lowest estimated expected costs over a simulation trajectory \citep{temizoz2023deep}. Algorithm~\ref{alg:DCL} provides a high-level overview of the \gls{DCL} training procedure. We initialise by setting a random initial policy, and several hyperparameters (1). Next, we conduct $n$ policy improvement steps (2). In each step, we initialise an empty data set (3), and start with a sampled starting state $s_0$ (4) and a set of sampled exogenous scenarios $\xi$ (5). Next, for each exogenous scenario (6), \gls{DCL} identifies the corresponding simulation-based decision $\hat{\pi}_i^+(\s_k)$ by locating the decision associated with the lowest estimate of the decision-value function (7). The found data point (state-decision pair) is added to the data set $\K_{i}$. Transitioning to a new state used for further data collection is done by employing the improved policy $\hat{\pi}_i^+(\s_k)$ (9). When enough data is collected, a classifier is trained on the collected data to obtain a new policy $\pi_{i+1}$ (10). After $n$ iterations, the found policies for each iteration are returned (11). For a more detailed overview of \gls{DCL}, we refer to \citet{temizoz2023deep}. We provide the same features to \gls{DCL} as we provide to the \gls{AVI} policy.

\newcommand{\Input}{\mathbf{\Xi}}
\begin{algorithm}[H]
\caption{{\gls{DCL}}}\label{alg:DCL}

\DontPrintSemicolon
{Initial policy:  $\pi_{0}$; neural network structure: $\NN_{\theta}$; number of approximate policy iterations: $n$; number of states to be collected: $N$; number of exogenous scenarios per state-decision pairs: $M$; depth of the rollouts (horizon length): $H$; length of the warmup period: $L$; number of workers: $w$}

\For{$i=0, 1, \dots, n-1 $}{
    $\K_{i} = \{\}$, the dataset\;
        $\s_0 = \text{SampleStartState}(\pi_i, L)$\;
        Generate exogenous scenario $\xi$, $|\xi| = {N }$\;
        \For{$k=1, \dots, {N }$}{
            Find $\hat{\pi}_i^+(\s_k) = \text{Simulator}(\s_{k},\pi_{i}, M, H)$\;
            Add ($\s_{k}$, $\hat{\pi}_i^+(\s_k)$) to the data set $\K_{i}$\;
            $\s_{k+1} = f(\s_{k}, \hat{\pi}_i^+(\s_k), \xi_{k-1})$\;
        }
    $\pi_{i+1} = \text{Classifier}(N_{\theta}, \K_i)$\;
}
{$\pi_{1}, \dots, \pi_{n}$}
\end{algorithm}

\section{{\gls{PPO}}}\label{app:PPO}

{\gls{PPO}} is a state-of-the-art actor-critic \gls{RL} framework, proposed by \citet{schulman2017proximal}. An actor-critic approach involves two main components: the actor, which proposes decisions given the current state, and the critic, which evaluates the proposed decisions. The \gls{PPO} algorithm improves upon traditional actor-critic methods by employing an objective function that balances exploration and exploitation. This is achieved through the use of a clipped probability ratio in the objective, which limits the updates to the policy in each iteration, preventing too drastic changes. Algorithm~\ref{alg:algo3} outlines the \gls{PPO} algorithm as detailed in \citet{schulman2017proximal}. We begin by initialising the network weights ${w}$ for the critic and ${\theta}$ for the actor (1), followed by setting hyperparameters such as the learning rates $\alpha_{\mathrm{cr}}$ and $\alpha_{\mathrm{ac}}$ for the critic and actor, respectively (2). After initialising a state ${s}_0$ (3), we enter a loop for each timestep in the horizon (4). A decision ${{a}}$ is generated by sampling from the policy $\pi_{{\theta}}$, which outputs probabilities for each decision, using a softmax output layer (5). Upon applying decision ${a}$ to the environment, we observe the transition to the subsequent state ${s}_{t+1}$ (6), after which we store the state, actions, and rewards in the trajectory buffer $\mathcal{T}$ (7). Every $T$ steps, updates are made to the actor and critic networks (9). Advantages are computed using the truncated Generalised Advantage Estimation (GAE) as follows (10):
\begin{equation}
\hat{A}_t(r,{s}_t,{{a}},{s}_{t+1}) = \sum_{t'=t}^{T}(\lambda\gamma)^{t'-t}\delta_{t'},
\end{equation}
where $\lambda$ is the temporal difference discount parameter and $\delta$ is defined by:
\begin{equation}
    \delta = r + \gamma\, Q({s}_{t+1},{{a}},{w}) - Q({s}_t,{{{a}}},{w}).
\end{equation}

We then proceed to optimise the policy loss (11) as described in \citet{schulman2017proximal} and the value loss (12) based on the $n$-step return:
\begin{equation}
    G_t^n = r_t + \gamma\,r_{t+1} + \ldots + \gamma^{n}\,Q_{{w}}({s}_{t+n}, {{a}}).
\end{equation}

The trajectory buffer $\mathcal{T}$ is emptied thereafter (13). For \gls{PPO}, we provide the same state features as we provide to \gls{AVI} and \gls{DCL}.

\begin{algorithm}[H]
\caption{{\gls{PPO}} algorithm }\label{alg:algo3}
\DontPrintSemicolon  % Disable the end-of-line semicolons

\SetKwInput{KwInput}{Input}
\SetKwInput{KwOutput}{Output}

Initialise critic and actor network weights ${w}$, ${\theta}$\;
Set hyperparameters: $\alpha_{\mathrm{cr}}$, $\alpha_{\mathrm{ac}}$\;
\For{each episode}{
    \For{$t=1,2,\ldots,T$}{
        ${{a}} \gets \pi_{{\theta}}({s}_t)$;
        Apply ${a}$ to environment, observe successor state ${s}_{t+1}$\;
        Store $({s}_t,{{a}},{s}_{t+1})_t$ in $\mathcal{T}$\;
    }
    Compute negative rewards $r$ and add them to $\mathcal{T}_T$\;
    Compute advantages $\hat{A}_t(r,{s}_t,{{a}},{s}_{t+1})$ and $\log\left(\pi_{{\theta}}({{a}})\right)$\;
    Optimise clipped policy loss based on $\hat{A}(r,{s}_t,{{a}},{s}_{t+1})$ and $\log\left(\pi_{{\theta}}({{a}})\right)$\;
    Optimise critic loss based on $n$-step return\;
    Empty $\mathcal{T}$\;     
}

\end{algorithm}

\section{Hyperparameters}\label{appendix:hyper}

In Table~\ref{tab:all_hyperparameters} we report the hyperparameters which we tested and eventually selected. Note that we only tune hyperparameters once for the synthetic case and for the energy case. We applied a search over the values in the \enquote*{Set of values} column and selected the values in the right columns.

The value for $S$, which bounds the decision space to accommodate for the exact benchmark, is set to $\{8,7,8,7,10,9,10,9,6,11\}$ in the 10 synthetic instances. For the energy case we set it to a fixed value of $25$.

For both \gls{AVI} and \gls{DCL}, $N$ indicates the number of samples that need to be collected, $H$ indicates the simulated horizon length. $\epsilon$ indicates the probability of taking a random decision during training, instead of on-policy transitions.

For models that were embedded in EPL, we increased the number of hidden layers by one (+1) for \gls{DCL} and \gls{PPO}. Furthermore, we increased $N$ by a factor $2.5$ and $M$ by a factor $2$. For AVI, we did not change hyperparameters when using EPL. We set the value of $\kappa_j$ to $10$ for all the energy case experiments, i.e., we create a grid with 10 values for each parameter $j$. For the synthetic case we do not use $\kappa_j$, as we sample the ten instances directly. Since we consider $12$ parameters for the the energy case (excluding $Var_A$ and $Var_C$ as demand follows a Poisson process), we have a grid $\Theta$ of size $10^{12}$.
\clearpage
\begin{table}[htbp]
	\centering
	 \def\arraystretch{1.1}
	\caption{Hyperparameter tuning.}
	\label{tab:all_hyperparameters}
	\resizebox{\textwidth}{!}{\begin{tabular}{llcc c}
		\toprule
   &  &  & \multicolumn{2}{c}{Chosen values} \\
        \cmidrule(r){4-5}
		& Hyperparameters & Set of values & Synthetic case & Energy case\\
		\midrule
  \multirow{3}{*}{\begin{sideways} \gls{AVI} \end{sideways}} & N & $\{2000, 4000\}$ & $2000$  & $2000$ \\
   & H  & $\{250,500\}$ & 250 & 500 \\
     & $\epsilon$ & $\{0.001,0.1,0.25\}$ & 0.1 & 0.1 \\
        \midrule
   \multirow{7}{*}{\begin{sideways} \gls{DCL} \end{sideways}} & N & $\{5000,10000, 20000,50000\}$ & $5000$ & $20000$  \\
   & M  & $\{1000,2000\}$ & 2000 & 2000 \\
    & H  & $\{100,500,200, 1000\}$ & 500 & 1000 \\
     & L  & $\{0,10,50\}$ & 10 & 10 \\
     & Hidden layers & $\{1,2,3\}$ & 2 & 3 \\
     & Nodes/layer & $\{16,32,64\}$ & 32 & 64\\
     & Batch size & $\{32,64,128, 256\}$ & 64 & 156  \\
  \midrule
       \multirow{8}{*}{\begin{sideways} \gls{PPO} \end{sideways}} & $\alpha^{\gls{PPO}}_{cr}$ (learning rate critic) & $\{10^{-2},10^{-3},10^{-4}\}$ & $10^{-3}$& $10^{-3}$ \\
        & $\alpha^{\gls{PPO}}_{ac}$ (learning rate actor) & $\{10^{-2},10^{-3},10^{-4}\}$ & $10^{-4}$ &  $10^{-4}$  \\
         & Hidden layers & $\{1,2,3\}$ & 3 & 3 \\
         & Nodes/layer critic & $\{16,32,64\}$ & 32 & 32 \\
         & Nodes/layer actor & $\{16,32,64\}$ & 64 & 64 \\
         & Batch size & $\{32,64,128\}$ & 128 & 128 \\
         & Clipping factor & $\{0.1,0.2,0.3\}$ & 0.2 & 0.2 \\
         & GAE $\lambda$ & $\{0.9,0.95,0.99,1.0\}$ & 0.95 & 0.95  \\
		\bottomrule
	\end{tabular}}
\end{table}

\newpage

\end{document}